\documentclass[journal]{IEEEtran}
\usepackage{amsmath,amsfonts}
\usepackage{algorithmic}
\usepackage{algorithm}
\usepackage{array}
\usepackage{multirow}
\usepackage{booktabs}
\usepackage{amsmath, amssymb}
\usepackage[caption=false,font=normalsize,labelfont=sf,textfont=sf]{subfig}
\usepackage{textcomp}
\usepackage{stfloats}
\usepackage{url}
\usepackage{verbatim}
\usepackage{graphicx}
\usepackage{cite}
\begin{document}

\title{Effective Gaussian Management for High-fidelity Scene Reconstruction}

\author{Jiateng Liu, Hao Gao, Jiu-Cheng Xie, Chi-Man Pun, Jian Xiong, Haolun Li, Junxin Chen, Feng Xu
\thanks{Jiateng Liu, Hao Gao, Jiu-Cheng Xie, and Haolun Li are with the School of Automation, Nanjing University of Posts and Telecommunications, Nanjing, 210023, China. E-mail: jiatliux@gmail.com, tsgaohao@gmail.com, jiuchengxie@gmail.com, lhl219319@gmail.com.}
\thanks{Jian Xiong is with the school of Communications and Information Engineering, Nanjing University of Posts and Telecommunications, Nanjing, 210023, China. E-mail: jxiong@njupt.edu.cn.}
\thanks{Junxin Chen is with the School of Software, Dalian University of Technology, Dalian 116621, China. E-mail: junxinchen@ieee.org.}
\thanks{Chi-Man Pun is with the Department of Computer and Information Science, University of Macau, Taipa, Macau. E-mail: cmpun@um.edu.mo.}
\thanks{Feng Xu is with the School of Software and BNRist, Tsinghua University, Beijing 100084, China. E-mail: xufeng2003@gmail.com.}
\thanks{Hao Gao and Feng Xu are the corresponding authors.}
}
\markboth{Journal of \LaTeX\ Class Files,~Vol.~14, No.~8, August~2021}%
{Shell \MakeLowercase{\textit{et al.}}: A Sample Article Using IEEEtran.cls for IEEE Journals}


\maketitle

\begin{abstract}
This paper proposes an effective Gaussian management framework for high-fidelity scene reconstruction of both appearance and geometry.
Unlike recent Gaussian Splatting (GS) pipelines that treat all primitives uniformly during optimization, our framework explicitly manages the attribute activation, representation and pruning of Gaussian.
Specifically, our framework first introduces \emph{GauSep}, a novel densification strategy that selectively activates Gaussian color or normal attributes to alleviate destructive gradient conflicts arising from dual supervision.
We further propose \emph{GauRep}, an adaptive Gaussian representation that dynamically adjusts spherical harmonics (SHs) orders and performs task-decoupled pruning to reduce redundancy at both the individual and global levels.
To provide reliable geometric supervision for above mangement process, we additionally introduce \emph{CoRe}, an regularized
surface reconstruction module that distills robust normal fields from an SDF branch to the Gaussian representation through a confidence mechanism.
Notably, the proposed Gaussian management is compatible with various reconstruction architectures and can be seamlessly integrated to improve performance while reducing size of the model.
Extensive experiments demonstrate that our approach achieves superior or comparable performance in appearance and geometry reconstruction compared with state-of-the-art methods, while using significantly fewer parameters.
\end{abstract}


\section{Introduction}
\label{sec:intro}
\IEEEPARstart{H}{igh-quality} scene reconstruction from a set of RGB images is critical for applications such as augmented/virtual reality (AR/VR), 3D content production, and edge computing \cite{yang2024deformable}. 
Although traditional methods \cite{izadi2011kinectfusion, pizzoli2014remode} have advanced substantially, they still struggle to handle complex scenes with robustness. 

Recently, neural implicit and explicit representations, most notably neural radiance fields (NeRF) and 3D Gaussian Splatting (3DGS), have advanced the state of the art in view synthesis and scene reconstruction. 
NeRF \cite{mildenhall2021nerf} models the scene as a continuous field parameterized by multi-layer perceptrons (MLPs), and NeuS \cite{wang2021neus} further introduces an SDF-derived density to improve geometric optimization. 
The follow-up works focus on improving the reconstruction fidelity and the training efficiency \cite{zhang2020nerf++, muller2022instant}. 
Although these works have significantly advanced the development of implicit neural fields, they remain computationally expensive due to the costly ray-marching required for rendering \cite{singh2024hdrsplat}.

By contrast, 3DGS represents the scene as a collection of unstructured splats and achieves real-time and high-quality rendering via GPU-accelerated rasterization \cite{zwicker2002ewa}. Nevertheless, the discrete, point-like nature of splats and the ambiguity of their geometric attributes pose challenges for accurate surface reconstruction. To mitigate these shortcomings, recent studies propose flattening splats to better define normals and introduce regularization schemes to improve surface fidelity \cite{huang20242d, guedon2024sugar, Dai2024GaussianSurfels}.

Despite these improvements, existing Gaussian splatting (GS) pipelines still suffer from fundamental limitations when jointly optimizing appearance and geometry.
In particular, most existing GS pipelines treat all Gaussians uniformly during optimization, without considering the different requirements imposed by appearance and geometry supervision. This design leads to two key challenges.

\begin{figure}[t]
    \centering
    \includegraphics[width=0.98\linewidth]{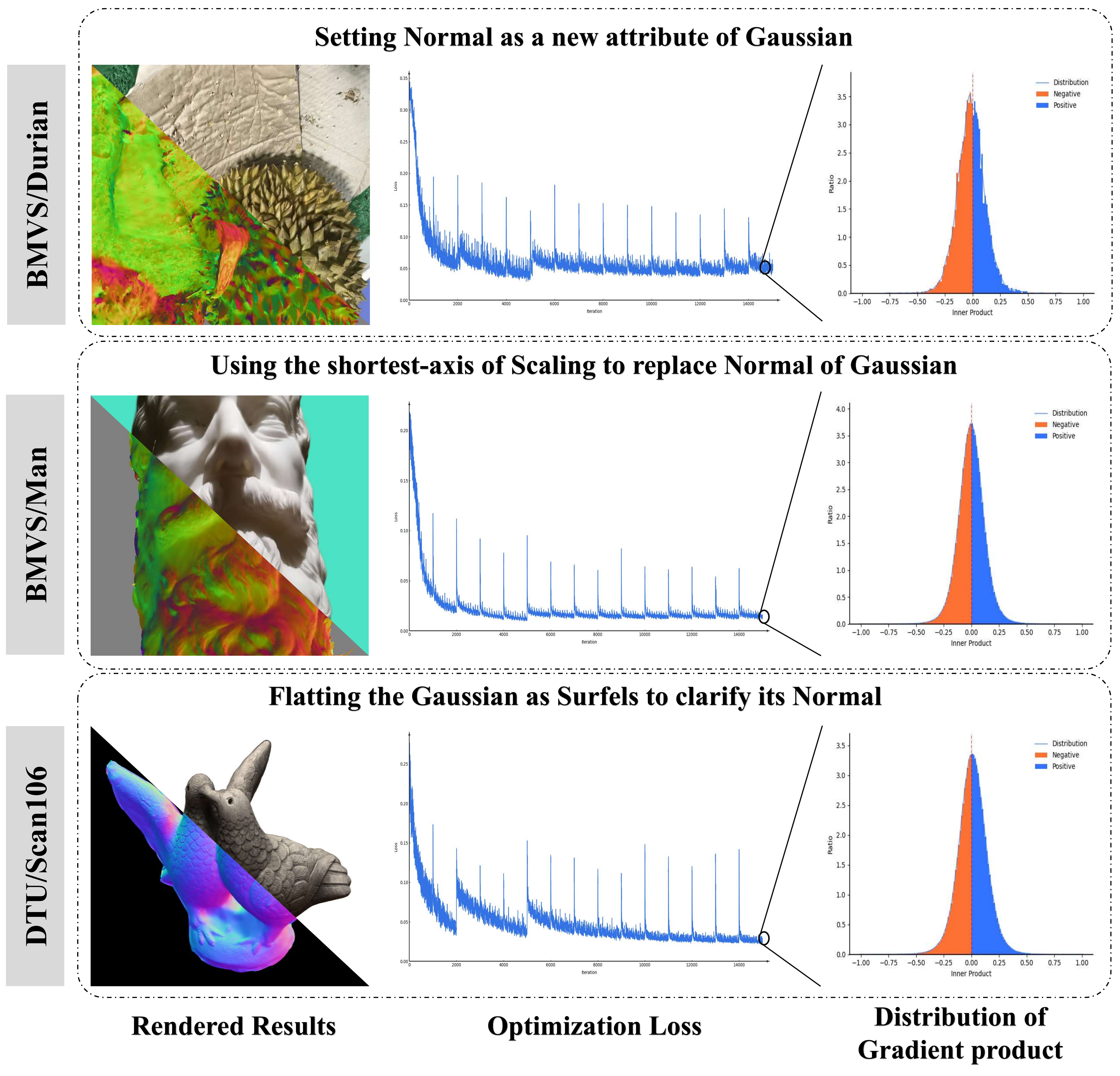}
    \caption{\textbf{Comparative analysis of gradient conflicts induced by different Gaussian-normal definitions.} The three columns show: (left) scene-fitting results, (middle) loss-optimization traces, and (right) the inner product between gradients of the RGB image and the normal map with respect to Gaussian attributes. For the inner product, the \emph{sign} denotes whether RGB- and normal-driven gradients are aligned (positive) or opposed (negative), and the \emph{magnitude} measures the severity of their disagreement. We evaluate three representative normal-definition strategies: (i) an explicit learnable normal attribute per Gaussian, (ii) alignment with the Gaussian's shortest principal axis, and (iii) deriving the normal from a flattened Gaussian (Surfels). All strategies exhibit gradient reversals during optimization; strategy (i) yields the most pronounced conflicts, whereas strategies (ii) and (iii) show similar patterns due to Gaussian flattening during optimization. These results underline the importance of conflict-aware Gaussian management when jointly supervising RGB and normal signals.}
\label{fig:gradient_conflicting}
\end{figure}

\emph{(1) Conflicting optimization.} The appearance and geometry supervision often produce conflicting gradients when applied to the same Gaussian attributes (e.g., position, etc.), which can hinder stable optimization and degrade reconstruction quality.
Fig.~\ref{fig:gradient_conflicting} illustrates the inner product between the gradients of Gaussian attributes obtained under the supervision of appearance and geometry, where each configuration (row) corresponds to a distinct method for Gaussian normal definition.

\emph{(2) Redundant representation.} Redundancy arises from both large number of Gaussian and high-order spherical harmonics (SHs). Standard densification strategies~\cite{kerbl3Dgaussians} often yield many Gaussians with negligible rendering contribution, which increases the representation size without commensurate benefit. Concurrently, applying a fixed high-order spherical-harmonic (SH) basis to every Gaussian forces a uniform expressive capacity regardless of local scene complexity, producing inefficiencies and representational waste.
\begin{figure*}[t]
    \centering
    \includegraphics[width=0.98\linewidth]{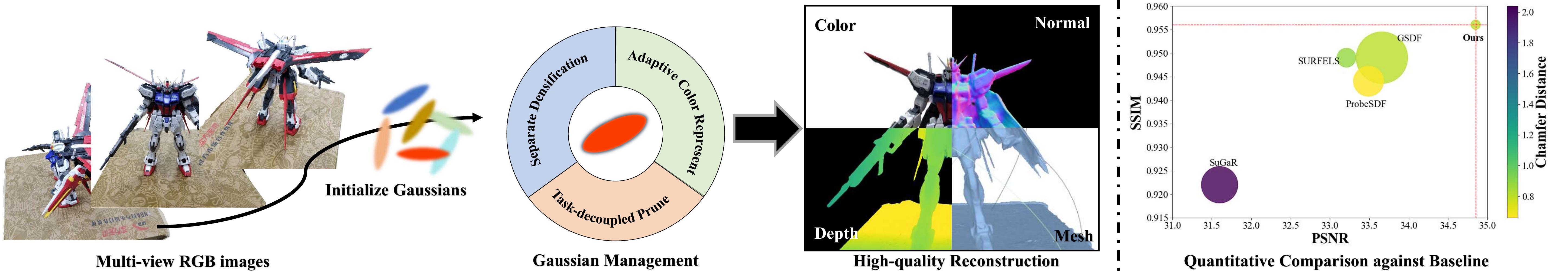}
    \caption{\textbf{Left}: Leveraging surface-aligned Gaussian surfels and the effective management strategies, our approach achieves comprehensive reconstruction of appearance and geometry from multi-view images. \textbf{Right}: The 2D plot shows that our approach achieves competitive results compared with the SOTA methods on the DTU \cite{aanaes2016large} dataset. The x-axis represents PSNR, and the y-axis represents SSIM. The circle color represents the Chamfer distance (CD, lower is better) for the reconstructed surface, while circle size represents the number of parameters (smaller is better).}
    \label{fig:teaser1}
\end{figure*}

Above issues indicate that simply optimizing a large set of Gaussian is insufficient. Therefore, we propose an adaptive and effective management framework to manage Gaussian attributes, representation capacity, and density for high-fidelity reconstruction. The framework combines an attribute-decoupled densification strategy, \emph{GauSep}, with an adaptive and integrated Gaussian representation, \emph{GauRep}. \emph{GauSep} monitors the gradient information during optimization and selectively activates the Gaussian color and normal attributes based on the correlation between their gradients. This design effectively alleviates multi-task gradient conflicts. Furthermore, \emph{GauRep} reduces redundancy at both individual and global levels. At the individual level, it adapts the spherical harmonics (SHs) order of each Gaussian based on its gradient magnitude, preserving representational capacity while reducing parameters. At the global level, it applies task-decoupled pruning to evaluate Gaussian contributions to appearance and geometry reconstruction separately, selectively removing those with the low task-specific contribution. Together, these components balance representational capacity with parameter efficiency while maintaining high reconstruction quality.

However, the effectiveness of such management strategies critically depends on accurate supervision, particularly geometric supervision. To establish this essential foundation, we introduce \emph{CoRe}, a surface reconstruction module based on GaussianSurfel~\cite{Dai2024GaussianSurfels}. We also propose a \emph{Confidence Mechanism} to distill robust normal fields from the Signed Distance Function (SDF) to complement the reconstruction of Gaussian, providing reliable geometric supervision for above Gaussian management.

The contributions of our work can be summarized as follows.
\begin{itemize}
    \item An attribute-decoupled densification strategy that enables selective attribute activation and thereby mitigate gradient conflicts during optimization.
    \item An adaptive and integrated Gaussian representation that reduces redundancy at both global and individual levels, balancing model capacity and parameter requirement.
    \item Compared with state-of-the-art methods, our approach achieves competitive or superior performance in appearance and geometry reconstruction while demonstrating robust compatibility.
\end{itemize}

\begin{figure*}[t]
\centering
\includegraphics[width=0.98\textwidth]{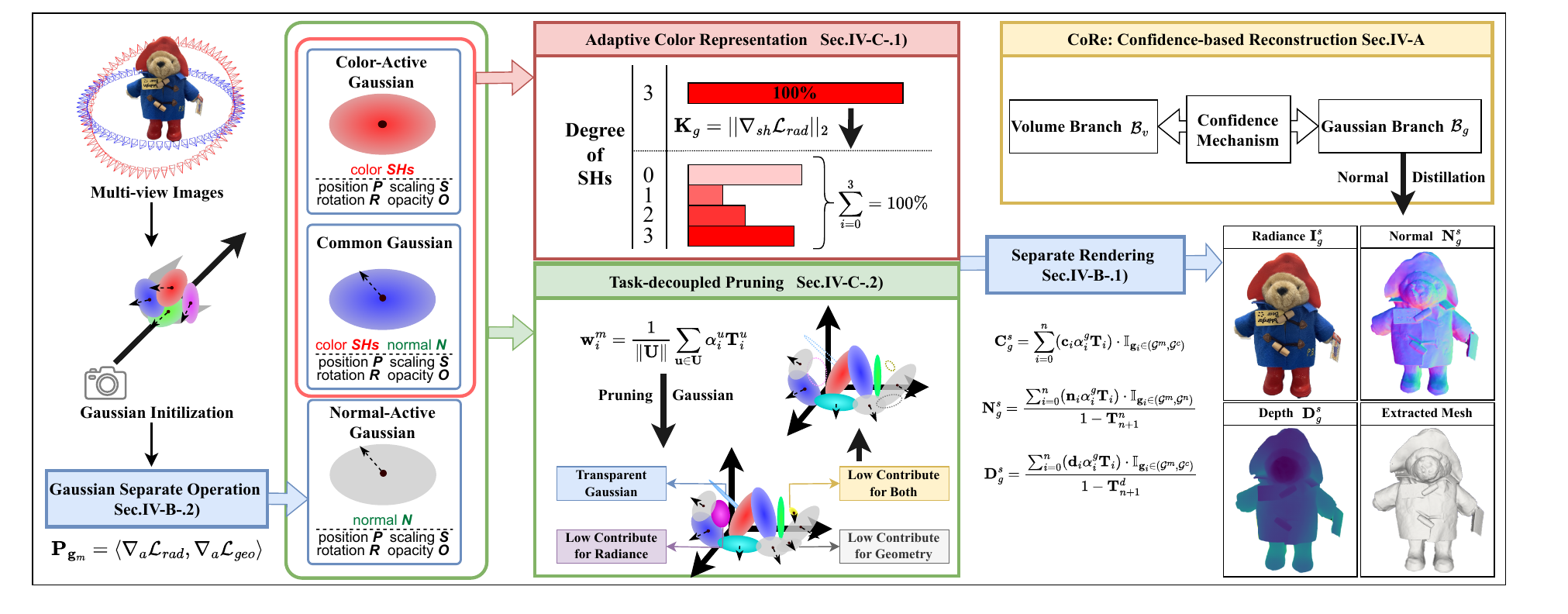}
\caption{\textbf{An illustration of our approach.} Given multi-view RGB images, we represent the scene using a collection of Gaussians, which are partitioned into three subsets through the \emph{Gaussian Separate Operation}: Common Gaussians, Color-active Gaussians, and Normal-active Gaussians. We further introduce \emph{GauRep}, an adaptive and integrated Gaussian representation, to reduce redundancy both at the individual level (\emph{adaptive color representation}) and the global level (\emph{task-decoupled pruning}). Task-decoupled pruning is applied to all Gaussian subsets, while adaptive color representation is specifically employed for those with active color attributes (i.e., Common Gaussians and Color-active Gaussians). To ensure high-quality normal supervision, we further propose \emph{CoRe}, a surface reconstruction module that distills normal fields from the SDF-based implicit branch $\mathcal{B}_{v}$ to the Gaussian branch $\mathcal{B}_{g}$ through a novel confidence mechanism. Finally, high-fidelity and competitive renderings, including images, normal maps, and depth maps, are produced through \emph{Separate Rendering}, which is tailored to the three Gaussian subsets. We employ screened Poisson reconstruction to extract surface from the normal and depth maps.}
\label{fig:pipeline}
\end{figure*}

\section{Related Work}
\label{sec:rel}
The reconstruction task focuses on both appearance and geometry, with the aim of producing representations that accurately capture objects or unbounded scenes. Early works \cite{cazals2006delaunay, seitz2006comparison} focus on exploring image features, leveraging multi-view consistency, and introducing prior knowledge to complete the reconstruction. However, due to ambiguities in the correspondence, the above methods cannot complete the task with high accuracy.
Recently, neural rendering methods have demonstrated powerful reconstruction capabilities by combining the powerful fitting ability of neural fields with a complete theoretical frameworks.

\subsection{Appearance Reconstruction and Rendering}
\label{sec:rel_app}
Neural fields were pioneered by NeRF \cite{mildenhall2021nerf}, which employs MLPs to model the radiance field for novel view synthesis. 
Subsequent work \cite{barron2022mip} reduces aliasing in unbounded 360 ° scenes through cone integration and distortion regularization, but remains reliant on computationally intensive deep MLPs. 
Instant-NGP \cite{muller2022instant} significantly accelerates the training time to seconds using multi-resolution hash coding. Despite these efficiency gains, the implicit representation limits the application and the structural compactness. In contrast, 3D Gaussian Splatting (3DGS) \cite{kerbl3Dgaussians} explicitly represents scenes with anisotropic Gaussian, enabling high-quality and real-time rendering with GPU/CUDA-based rasterization, but exhibits high redundancy due to the dense Gaussians needed to fit the details of the scene. Scaffold-GS \cite{lu2024scaffold} mitigates this issue through a structured Gaussian representation guided by sparse anchors, where compact MLPs predict local Gaussian attributes from sparse anchor points, thereby reducing redundancy and improving efficiency.
Spec-Gaussian \cite{yang2024spec} extends 3DGS with a dual-branch design for diffuse/specular components, enhancing reflective scenes via view-dependent directional modulation. However, these methods lack geometric consideration, impacting scalability and practicality. In contrast, our approach achieves high-fidelity reconstruction of both appearance and geometry simultaneously.

\subsection{Surface Reconstruction and Representation}
\label{sec:rel_sur}
Although radiance field methods have advanced scene reconstruction and novel view synthesis, they rely on heuristic density thresholds for surface extraction \cite{Dai2024GaussianSurfels}, often introducing noise and artifacts. Consequently, more efficient SDF-based methods gradually replace them, producing higher-fidelity surfaces. NeuS~\cite{wang2021neus} models surfaces using an SDF-derived density distribution. Subsequent work \cite{wang2023neus2, li2023neuralangelo} incorporates second-order/higher-order derivatives to improve surface fidelity.
However, these methods have significant rendering costs. Recent point-based approaches focus on 3DGS \cite{guedon2024sugar, yu2024gsdf, Dai2024GaussianSurfels}. SuGaR \cite{guedon2024sugar} regularizes splats for surface alignment, enabling efficient extraction. NeuSG \cite{chen2023neusg} and GSDF \cite{yu2024gsdf} integrate splats with SDF for improved results, but their SDF branch compromises efficiency. GaussianSurfels \cite{Dai2024GaussianSurfels} flattens splats into unambiguous surfels and uses screened Poisson reconstruction \cite{kazhdan2013screened} for meshing instead of direct extraction. Although this approach yields a high-fidelity surface, it loses accuracy in reconstructing the appearance, whereas our approach alleviates the problem by effective management of Gaussian attribute.

\section{Preliminaries}
\subsection{Neural Implicit Representations}
NeRF \cite{mildenhall2021nerf} is a seminal neural implicit representation. Subsequent work \cite{wang2021neus} integrates a signed distance function (SDF) into the NeRF pipeline by converting the SDF values at sample points $\mathbf{x}_{i}$ into alpha weights $\alpha_{i}^{v} $:
\begin{equation}
\label{eq:sdf2alpha}
\alpha_{i}^{v} = \max(\frac{\sigma(\mathcal{S}(\mathbf{x}_{i}))-\sigma(\mathcal{S}( \mathbf{x}_{i+1}))}{\sigma(\mathcal{S}(\mathbf{x}_{i}))}, 0),
\end{equation}
where $\sigma(\cdot)$ denotes the sigmoid function and $\mathcal{S}(\cdot)$ denotes the SDF modeled by an MLP. The pixel color $\mathbf{C}_{v}$ is obtained by integrating the radiance outputs $\mathbf{c}^{v}$ of the radiance MLPs $\Theta_{c}$ along the viewing ray.
\begin{equation}
\label{eq:volume_rendering}
\mathbf{C}_{v}(\mathbf{r}) = \sum^{n}_{i=1} \mathbf{c}^{v}_{i}\, \alpha^{v}_{i}\, \mathbf{T}^{v}_{i}\ ,
\quad
\mathbf{T}_{i}^{v} = \exp \lbrace - \sum_{j=1}^{i-1}{\alpha_j^{v}\, \delta_j} \rbrace,
\end{equation}
where $ \mathbf{T}^{v} $ and $ \delta $ denote the transmittance and the interval of the sample points, respectively. 

\subsection{Gaussian Surfels}
Following the idea of flattening splats to clarify geometric properties \cite{Dai2024GaussianSurfels}, we parameterize each Gaussian $\mathbf{g}_{i}$ by its center $\mathbf{p}_{i}$, rotation $\mathbf{R}_{i}= \left[\mathbf{R}_{i}^{x}, \mathbf{R}_{i}^{y}, \mathbf{R}_{i}^{z}\right]$, scaling $\mathbf{S}_{i} = \left[s_{i}^{x}, s_{i}^{y}, 0 \right]^{\mathrm{T}}$, opacity $\mathbf{o}_{i}$, normal $\mathbf{n}_{i} = \mathbf{R}_{i}^{z}$ and directional color $\mathbf{c}_{i}$ (represented via spherical harmonics, SHs), where $i$ indexes Gaussians in the scene. The covariance is then given by:
\begin{equation}
\label{eq:gaussian_covariance}
\Sigma = \mathbf{R}\mathbf{S}\mathbf{S}^{\mathrm{T}}\mathbf{R}^{\mathrm{T}} = \mathbf{R}\mathbf{Diag}\left[(s_{i}^{x})^{2}, (s_{i}^{y})^{2}, 0 \right] \mathbf{R}^{\mathrm{T}},
\end{equation}
where $\mathbf{Diag}[\cdot]$ denotes the diagonal matrix with diagonal entries in $\left[\right]$. Therefore, the Gaussian distribution of the pixel $\mathbf{u}$ can be represented as follows.
\begin{equation}
\label{eq:gaussian_distribution}
\mathbf{G}(\mathbf{u}; \mathbf{p}_{i}, \Sigma_{i}) = \exp \lbrace - 0.5 (\mathbf{u} - \mathbf{p}_{i})^{T} \Sigma_{i}^{-1} (\mathbf{u} - \mathbf{p}_{i}) \rbrace,
\end{equation}
We further leverage the affine approximation of the projective transformation $\mathbf{J}_{I}$ and the viewing transformation matrix $\mathbf{W}_{I}$ of the image $\mathbf{I}$ to project the Gaussian onto the image plane. The weight of Gaussian $\alpha_{i}^{g}$ is calculated follows.
\begin{equation}
\label{eq:gaussian_weight}
\alpha^{g}_{i} = \mathbf{o}_{i} \mathcal{P}\lbrace\mathbf{G}(\mathbf{u}; \mathbf{p}_{i}, \Sigma_{i}); \mathbf{J}_{I}, \mathbf{W}_{I}\rbrace,
\end{equation}
where $\mathcal{P}(\cdot)$ is the projection operator introduced in \cite{zwicker2002ewa}. Finally, the color $\mathbf{C}_{g}$ of the pixel $\mathbf{u}$ is rendered as:
\begin{equation}
\label{eq:gaussian_splatting}
\mathbf{C}_{g} = \sum^{n}_{i=0}\mathbf{c}^{g}_{i}\alpha^{g}_{i}\mathbf{T}^{g}_{i}, \quad \mathbf{T}^{g}_{i} = \prod^{i-1}_{j=0}(1 - \alpha^{g}_{j}),
\end{equation}
Eq.~\eqref{eq:gaussian_splatting} can be extended to render depth $\mathbf{D}_{g}$ or normals $\mathbf{N}_{g}$ by adding the appropriate normalization.
\cite{Dai2024GaussianSurfels} further introduces a geometry loss $\mathcal{L}_{geo}$ and a consistency loss $\mathcal{L}_{con}$ to supervise geometry and enforce spatial consistency.
\begin{equation}
\label{loss:surfel_geometry}
\mathcal{L}_{geo} = \lambda_{n}\mathcal{L}_{cos}(\mathbf{N}_{g}, \mathbf{N})\odot\mathbf{M},
\end{equation}
\begin{equation}
\label{loss:surfel_consistency}
\mathcal{L}_{con} = \lambda_{s}\mathcal{L}_{cos}(\mathbf{N_{g}, \mathcal{F}(\mathbf{D}_{g})})\odot\mathbf{M},
\end{equation}
where $\mathcal{F}(\cdot)$ converts depth to a normal map, $\odot$ denotes element-wise multiplication, and $\mathbf{N}$ is a normal supervision map produced by a pre-trained normal estimator (we updated it in Eq.~\eqref{eq:normal_update}). The ground truth mask $\mathbf{M}$ is applied to focus optimization on valid regions.
\begin{figure}[t]
\centering
\includegraphics[width=0.98\linewidth]{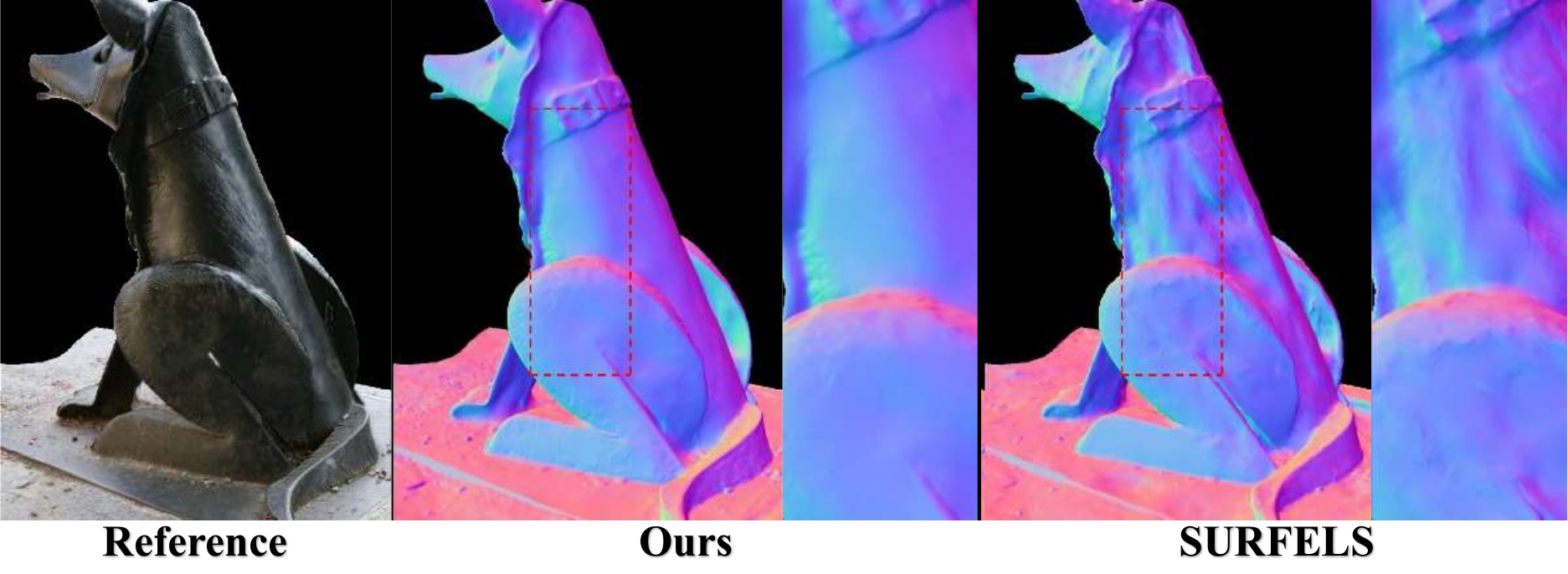}
\caption{\textbf{Effect of SDF Supervision.} Without SDF supervision, GaussianSurfels\cite{Dai2024GaussianSurfels} erroneously interpret specular highlights as geometric variations, leading to degraded surface quality. Incorporating SDF supervision effectively mitigates this issue and yields more faithful geometry reconstruction.}
\label{fig:surfel_fault}
\end{figure}
\section{Approach}
Given a set of RGB images of a target scene captured by calibrated cameras, our goal is to reconstruct both its appearance and geometry. 
We first introduce a surface reconstruction module, \emph{CoRe}, which jointly optimizes an SDF-based implicit branch $\mathcal{B}_{v}$ and a Gaussian-based representation branch $\mathcal{B}_{g}$. 
A confidence mechanism is employed to distill the normal field from $\mathcal{B}_{v}$ into $\mathcal{B}_{g}$ (Sec.~\ref{sec:core}). 
Subsequently, we use the CoRe-derived normal map as regularized geometry supervision and utilize the confidence-free Gaussian branch of $\mathcal{B}_{g}$ to perform effective and adaptive Gaussian management. 
This management comprises \emph{GauSep}, an attribute-decoupled densification strategy with separate rendering to alleviate gradient conflicts arising from dual supervision (Sec.~\ref{sec:gausep}), and \emph{GauRep}, an adaptive and integrated Gaussian representation to balance the representation capacity and model size (Sec.~\ref{sec:gaurep}). 
Fig.~\ref{fig:pipeline} illustrates our framework pipeline, which integrates \emph{CoRe} as a geometry regularizer to provide supervision, \emph{GauSep} for conflict-aware densification, and \emph{GauRep} for adaptive representation, enabling high-fidelity reconstruction with compact model size.

\subsection{CoRe: Confidence-Based Reconstruction Module}
\label{sec:core}
Although GaussianSurfel \cite{Dai2024GaussianSurfels} has demonstrated the strong capability for joint appearance and geometry reconstruction, its discrete representation fails to handle strong view-dependent effects such as specular highlights, as illustrated in Fig.~\ref{fig:surfel_fault} (right). Motivated by \cite{yu2024gsdf, wu2025deferredgs}, we augment the Gaussian representation branch $\mathcal{B}_{g}$ with an auxiliary SDF-based implicit branch $\mathcal{B}_{v}$ and couple them via a confidence mechanism to form a dual-branch architecture.

\subsubsection{Confidence Mechanism}
To leverage the complementary strengths of both representations, we define the confidence fields $f_{g}$ and $f_{v}$ for the Gaussian branch $ \mathcal{B}_{g} $ and the SDF branch $ \mathcal{B}_{v} $ as proxies for reconstruction quality, respectively.

In the Gaussian branch $ \mathcal{B}_{g} $, the confidence field $f_g$ is implemented by endowing each Gaussian with a scalar attribute $ \mathbf{f}^{g} \in (0, 1)$ that represents the significance level of each Gaussian during training.
The attribute is incorporated into the differentiable rasterization pipeline \cite{kerbl3Dgaussians} and produces a rendered significance map $ \mathbf{F}_{g} $ through the same composition in Eq.~\eqref{eq:gaussian_splatting}. 
\begin{equation}
\label{eq:gaussian_confidence_splatting}
\mathbf{F}_{g} = \sum^{n}_{i=1} \mathbf{f}^{g}_{i} \alpha_{i}^{g} \mathbf{T}_{i}^{g},
\end{equation}
Notably, the higher value of $ \mathbf{F}_{g} $ corresponds to the lower confidence in the reliability of the pixel.

In the SDF branch $ \mathcal{B}_{v} $, the confidence field is parameterized by appending a lightweight confidence MLPs $ \Theta_{f} $ alongside the radiance MLPs $ \Theta_{c} $. $ \Theta_{f} $ accepts the same inputs as $ \Theta_{c} $ but uses approximately half the parameters, which outputs confidence values $ \mathbf{f}_{v} $ of each sample points. Therefore, the confidence map in SDF branch is yielded by accumulated $ \mathbf{f}_{v} $ along rays according to Eq.~\eqref{eq:volume_rendering}. 
\begin{equation}
\label{eq:volume_confidence_rendering}
\mathbf{F}_{v}(r) = \sum^{n}_{i=1} \mathbf{f}^{v}_{i} \alpha_{i}^{g} \mathbf{T}_{i}^{g},
\end{equation}
Both $\mathbf{f}^{g}$ and $\mathbf{f}^{v}$ are motivated by a sigmoid function to constrain the values to $ (0,1) $.

Finally, we utilize the significance map $ \mathbf{F}_{g} $ in the Gaussian branch to modulate the consistency loss $\mathcal{L}_{con}$ to suppress excessive appearance-driven intervention in geometry reconstruction. The updated consistency loss is defined as follows.
\begin{equation}
\label{loss:adaptive_consistency_geometry}
\mathcal{L}_{con}^{a} = \mathcal{L}_{con} + \lambda_{s}\mathcal{L}_{cos}(\mathcal{F}(\mathbf{D}_{g}), \mathbf{N})\odot\mathbf{M}^{\prime}
\end{equation}

For the supervised normal $\mathbf{N}$ and its corresponding confidence-modulated mask $\mathbf{M}^{\prime}$, we adopt a two-stage strategy.
Initially, we use the prior normal $\mathbf{N}_{c}$ from a pretrained estimator \cite{eftekhar2021omnidata} with the Gaussian significance map $\mathbf{F}_{g}$ modulating the mask $\mathbf{M}$ to ensure stable convergence. After iteration $\mathcal{T}_{i}$, we switch to the SDF-derived normal $\mathbf{N}_{v}$ with combined confidence map $\mathbf{F}_{g}$ and $\mathbf{F}_{v}$, which provides more accurate geometric guidance distilled from the volumetric branch. This can be formulated as:
\begin{equation*}
\mathbf{N}, \mathbf{M}^{\prime}
=
\begin{cases}
\mathbf{N}_{c}, \mathbf{M}\odot\mathbf{F}_{g} &{\text{if}}\ iter \leq \mathcal{T}_{i}\\
\mathbf{N}_{v}, \mathbf{M}\odot\mathbf{F}_{g}\odot\mathbf{F}_{v} &\ \text{otherwise} \\
\end{cases}
,
\label{eq:normal_update}
\end{equation*}
where the value of $\mathcal{T}_{i}$ is discussed in Sec~\ref{sec:opt_training_details}.
\subsubsection{Confidence Optimization}
Rather than optimizing confidence related parameters directly with an unconstrained objective, we supervise the confidence fields using reconstruction metrics to avoid ambiguity. Specifically, we optimize $\mathbf{f}^{g}$ so that the rendered map $\mathbf{F}_{g}$ highlights pixels exhibiting low radiance error but high geometry error; the confidence network $\Theta_{f}$ is optimized to predict per-ray confidence based on appearance reconstruction quality and a geometric uncertainty measure (Shannon entropy) of the ray.

\paragraph{The attribute $\mathbf{f}^{g}$ in $ \mathcal{B}_{g}$}
We first define the radiance loss $ \mathcal{L}_{rad} $.
\begin{equation}
\label{loss:core_radiance}
    \mathcal{L}_{rad} = (1 - \lambda_{s}) \mathcal{L}_{1}(\mathbf{C}_{g}, \mathbf{C}) + \lambda_{s}\mathcal{L}_{ssim}(\mathbf{C}_{g}, \mathbf{C}), \\
\end{equation}
with $\lambda_{ssim}=0.2$ in all experiments and the geometry loss have defined in Eq.\eqref{loss:surfel_geometry}. The rendered significance map is supervised by a binary target map $\mathbf{F}_{g}^{gt}$ that identifies pixels where the radiance gradient magnitude is small while the geometry gradient magnitude is large:
\begin{align}
\label{loss:gaussian_confidence}
    \mathcal{L}_{\mathbf{F}_{g}} & = \vert\vert \mathbf{F}_{g} - \mathbf{F}_{g}^{gt} \vert\vert_{2}, \\
    \mathbf{F}_{g}^{gt} = \mathbb{I}(\vert\vert \nabla_p \mathcal{L}_{rad}\vert\vert& < \zeta_{rad} \And \vert\vert \nabla_R \mathcal{L}_{geo}\vert\vert) > \zeta_{geo}
\end{align}
where the $\nabla_p\mathcal{L}_{rad}$ and the $\nabla_R\mathcal{L}_{geo}$ denotes the partial derivatives of $\mathcal{L}_{rad}$ and $\mathcal{L}_{geo}$ with respect to the Gaussian positions and the rotations, respectively. We use $ \zeta_{rad} = 0.0002 $ and $ \zeta_{geo} = 0.0001 $ in all experiments.

\paragraph{The confidence network in $ \mathcal{B}_{v} $}
For the SDF branch, we define an appearance-based signal,
\begin{equation}
\label{eq:volume_confidence}
\mathbf{E}(r) = 1 - \| \mathbf{C}_{v}(r) - \mathbf{C}(r) \|_{1},
\end{equation}
where $\mathbf{C}(r)$ are the pixels of the ground truth RGB image used in the SDF branch $\mathcal{B}_{v}$. 
$\mathbf{E} \in [0,1]$ since the the L1 norm of color differences is bounded by 1 for normalized RGB values. We also define a geometric uncertainty measured by the Shannon entropy:
\begin{equation}
\label{eq:shannon_entropy}
\mathbf{H}(r) = 1 - \sum_{j=0}^{n} -\mathbf{h}_{j}\log(\mathbf{h}_{j}),\quad
\mathbf{h}_{j} = \frac{\alpha_{j}^{v}}{\sum_{i=0}^{n}\alpha_{i}^{v}}.
\end{equation}
The confidence network is trained to match both signals:
\begin{equation}
\label{loss:volume_confidence}
\mathcal{L}_{\mathbf{F}_{v}}(r) = \big\|\mathbf{F}_{v}(r) - \mathbf{E}(r)\big\|_{2} + \lambda_{H}\big\|\mathbf{F}_{v}(r) - \mathbf{H}(r)\big\|_{2},
\end{equation}
where $\lambda_{H}=5\times 10^{-3}$ in all experiments.

After training \emph{CoRe}, high-fidelity normal supervision $\mathbf{N}_{g}$ is obtained from the Gaussian branch $\mathcal{B}_{g}$. $\mathbf{N}_{g}$ leveraged as additional normal supervision for Gaussian management, specifically for attribute-decoupled densification in \emph{GauSep} \ref{sec:gausep}. Since $\mathbf{N}_{g}$ already encodes robust geometric priors, the subsequent management stage bypasses the need for running-time confidence computation.

\subsection{GauSep: Attribute-Decoupled Densification with Separate Rendering}
\label{sec:gausep}
To mitigate the gradient conflict inherent in GS-based methods and the varying requirements for the quantity of Gaussian across different supervision signals, we propose \emph{GauSep}, an attribute-decoupled densification strategy named \emph{Gaussian Separate Operation} and its associated rendering pipeline \emph{Separate Rendering}.
Our approach explicitly disentangles task-specific attributes, allowing each Gaussian to participate selectively in appearance or geometry reconstruction,
thereby mitigating destructive gradient interference and enabling more precise Gaussian management.
\begin{figure}[t]
    \centering
    \includegraphics[width=0.98\linewidth]{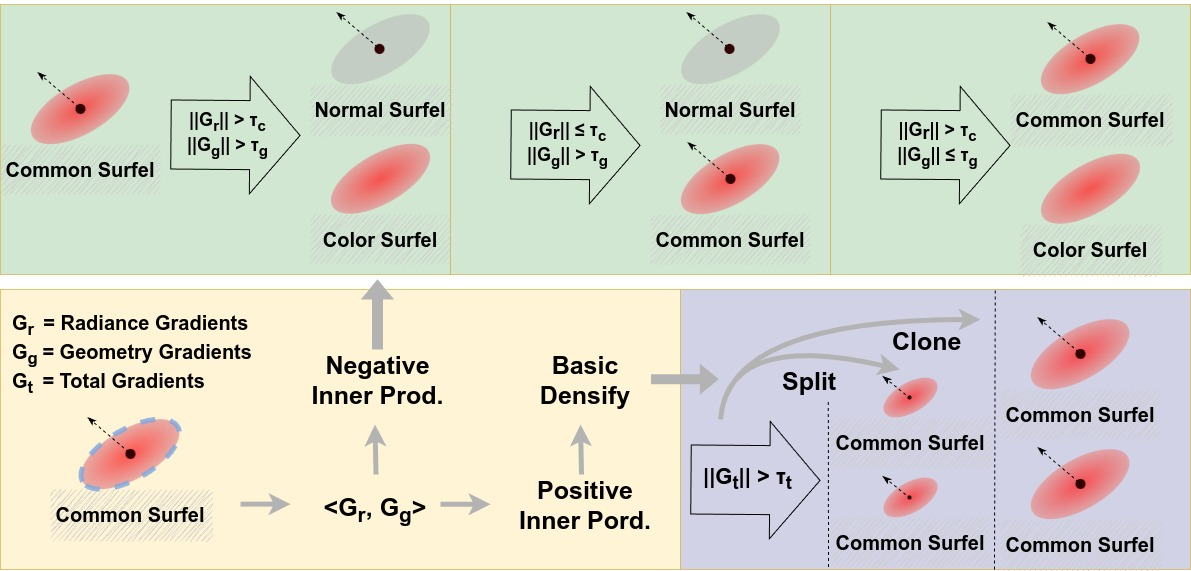}
    \caption{\textbf{Illustration of Gaussian Separate Operation.} For each Gaussian, we first compute the inner product of its radiance and geometry gradients. When the product is positive, standard clone and split operations are performed; otherwise, Gaussian Separate Operation applies attribute decoupling and creates separate Gaussians specialized for color or geometry. This results in three disjoint Gaussian sets that are optimized under distinct supervision signals for subsequent rendering.}
    \label{fig:separate_rendering}
\end{figure}

\subsubsection{Separate Rendering}
\begin{figure}[t]
    \centering
    \includegraphics[width=0.98\linewidth]{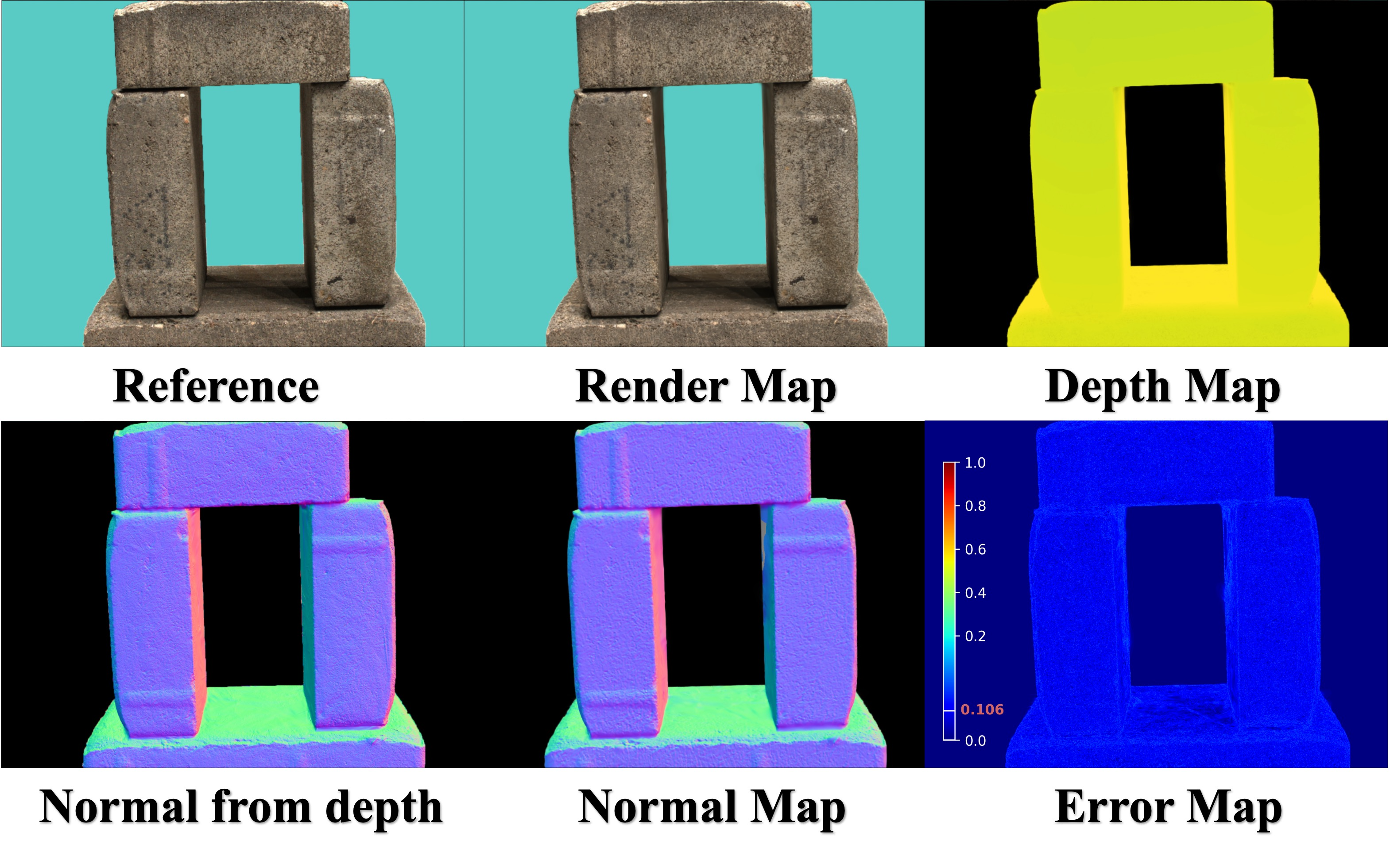}
    \caption{\textbf{Consistency Check of Gaussian Separate Operation.} \textbf{Top}: reference image, rendered appearance, and depth map. \textbf{Bottom}: normals computed from rendered depth, rendered normal map, and error map between two normal map by rendering or calculating from depth map. The rendered results illustrates high consistency cross appearance and geometry and the max error value is marked in red in the color bar of error map.}
    \label{fig:consistency_check}
\end{figure}

We update the attributes of each Gaussian and decompose them into basic attributes $ \lbrace \mathbf{p}, \mathbf{R}, \mathbf{S}, \mathbf{o} \rbrace $ and task-specific attributes $ \lbrace \mathbf{c}, \mathbf{n} \rbrace $. 
This decoupling allows Gaussians to selectively activate only color or normal attributes, producing three disjoint subsets:
the common set $\mathcal{G}^{m} = \{\mathbf{g}_{i} | \mathcal{A}(\mathbf{g}_{i}) = (\mathbf{p}, \mathbf{R}, \mathbf{S}, \mathbf{o}, \mathbf{n}, \mathbf{c})\}$,
the color-active set $\mathcal{G}^{c} = \{\mathbf{g}_{i} | \mathcal{A}(\mathbf{g}_{i}) = (\mathbf{p}, \mathbf{R}, \mathbf{S}, \mathbf{o}, \mathbf{c})\}$,
and the normal-active set $\mathcal{G}^{n} = \{\mathbf{g}_{i} | \mathcal{A}(\mathbf{g}_{i}) = (\mathbf{p}, \mathbf{R}, \mathbf{S}, \mathbf{o}, \mathbf{n})\}$,
where $\mathcal{A}(\mathbf{g})$ denotes the activated attributes of the Gaussian $\mathbf{g}$.
We further modify the rendering equation in GaussianSurfels \cite{Dai2024GaussianSurfels} by introducing an indicator function $\mathbb{I}$. This ensures that each Gaussian only participates in its corresponding rendering process.
This yields the rendered color map $\mathbf{C}_{g}^{s}$, depth map $\mathbf{D}_{g}^{s}$, and normal map $\mathbf{N}_{g}^{s}$ as:
\begin{equation}
\label{eq:separate_rendering_color}
\mathbf{C}_{g}^{s} = \sum_{i = 0}^{n} (\mathbf{c}_{i} \alpha_{i}^{g} \mathbf{T}_{i}) \cdot \mathbb{I}_{\mathbf{g}_{i} \in (\mathcal{G}^{m}, \mathcal{G}^{c})},
\end{equation}
\begin{equation}
\label{eq:separate_rendering_depth}
\mathrm{D}_{g}^{s} = \frac{1}{1 - \mathbf{T}^{d}_{n+1}}\sum_{i=0}^{n} (\mathrm{d}_{i} \alpha_{i}^{g} \mathbf{T}_{i}) \cdot \mathbb{I}_{\mathbf{g}_{i} \in (\mathcal{G}^{m}, \mathcal{G}^{c})},
\end{equation}
\begin{equation}
\label{eq:separate_rendering_normal}
\mathbf{N}_{g}^{s} = \frac{1}{1 - \mathbf{T}^{n}_{n+1}}\sum_{i=0}^{n} (\mathbf{n}_{i} \alpha_{i}^{g} \mathbf{T}_{i}) \cdot \mathbb{I}_{\mathbf{g}_{i} \in (\mathcal{G}^{m}, \mathcal{G}^{n})},  
\end{equation}
where $\mathrm{d}$ denotes the depth of each Gaussian, as defined in \cite{Dai2024GaussianSurfels}, and $i$ represents the index of the Gaussian that participates in the rendering process of a certain pixel.
This formulation preserves appearance fidelity while leveraging geometric priors through the consistency loss defined in Eq.~\eqref{loss:adaptive_consistency_geometry}.
Fig.~\ref{fig:consistency_check} also presents visualization results generated using the above rendering pipeline, demonstrating the geometric consistency among the rendered outputs.

\subsubsection{Gaussian Separate Operation}
We now detail the derivation of the three Gaussian sets.
The common set $\mathcal{G}_{m}$ serves as the foundation for assigning Gaussians to the color-active set $\mathcal{G}_{c}$ or the normal-active set $\mathcal{G}_{n}$ through gradient magnitude-guided transformations of Gaussian attributes.
Specifically, we record the gradients of each common Gaussian $\mathbf{g}_{m}$ with respect to the radiance loss $\mathcal{L}_{rad}$ and the geometry loss $\mathcal{L}_{geo}$ during optimization.
The degree of conflict between these signals is quantified by the inner product of their gradients:
\begin{equation}
\label{eq:separate_operation}
    \mathbf{P}_{\mathbf{g}_{m}} = \left< \nabla_a \mathcal{L}_{rad}, \nabla_a \mathcal{L}_{geo} \right> = \frac{\nabla_{a} \mathcal{L}_{rad} \cdot \nabla_{a} \mathcal{L}_{geo}}{\vert\vert \nabla_{a} \mathcal{L}_{rad} \vert\vert \cdot \vert\vert \nabla_{a} \mathcal{L}_{geo} \vert\vert},
\end{equation}
where $\nabla_a\mathcal{L}$ denotes the gradient of Gaussian attributes with respect to the loss function $\mathcal{L}$. We concatenate the gradients of all attributes and subsequently flatten them into a vector.
\begin{equation}
\label{eq:grad}
\nabla_a\mathcal{L} = \operatorname{Concat}\left(\left\{\nabla_{a_i}\mathcal{L}\right\}_{a_i \in \{\mathbf{p},\mathbf{R},\mathbf{S},\mathbf{o}\}}\right),
\end{equation}
where $\operatorname{Concat}(\cdot)$ reshapes each attribute gradient into a one-dimensional vector and concatenates them along the feature dimension, e.g., $\mathbb{R}^3 \oplus \mathbb{R}^4 \oplus \mathbb{R}^2 \oplus \mathbb{R}^1 \rightarrow \mathbb{R}^{10}$.

A Gaussian with negative product ($ \mathbf{P}_{\mathbf{g}_{m}} < 0 $) is separated into two new Gaussians with attributes adjusted according to the average magnitude of each gradient. 
Specifically, a common Gaussian that satisfies $ \vert\vert \nabla_{a} \mathcal{L}_{rad} \vert\vert > \tau_c \And \vert\vert \nabla_{a} \mathcal{L}_{geo} \vert\vert \leq \tau_g $ is separated into a common Gaussian and a color-active Gaussian, and vice versa.
a common Gaussian that satisfies $ \vert\vert \nabla_{a} \mathcal{L}_{rad} \vert\vert > \tau_c \And \vert\vert \nabla_{a} \mathcal{L}_{geo} \vert\vert > \tau_g $ is separated into a color-active Gaussian and a normal-active Gaussian.
The remaining Gaussians continue to perform the standard split and clone operation.
The specific operation steps are shown in Fig.~\ref{fig:separate_rendering}. The above operation is performed every $N$ iteration, where $N$ is the number of images.

Finally, the scene is represented by three Gaussian subsets, each contributing to the rendering results through its respective rendering pipeline (Eq.\eqref{eq:separate_rendering_color}, \eqref{eq:separate_rendering_depth}, \eqref{eq:separate_rendering_normal}).
Despite the effective representation, the scene still exhibits considerable redundancy.

\subsection{GauRep: Adaptive and Integrated Gaussian Representation}
\label{sec:gaurep}
In this section, we propose GauRep, an adaptive and integrated Gaussian representation to mitigate dual redundancy in GS-based methods. The \emph{global redundancy} caused by an excessive number of Gaussians and the \emph{individual redundancy} arising from inefficient fixed-order spherical harmonics (SHs). Our approach jointly leverages two complementary techniques, \emph{adaptive color representation} for reducing individual redundancy and \emph{task-decoupled pruning} for alleviating global redundancy.

\subsubsection{Individual, Adaptive Color Representation}
Existing methods \cite{fan2024lightgaussian, liu2024lgs} typically rely on SH distillation, which explicitly fixes the expressive capacity of each Gaussian. In contrast, our key innovation is to grant each Gaussian the flexibility to progressively and independently increase its SH order during optimization, thereby adapting its expressive power to the underlying scene content.

Concretely, all Gaussians are initialized with zeroth-order SHs, corresponding to three coefficients for directional color. 
During optimization, we accumulate the partial derivatives of the radiance loss $ \mathcal{L}_{rad} $ with respect to the SH coefficients. The magnitude of these accumulated gradients serves as the criterion for determining whether the order of the Gaussian should be increased.
\begin{equation}
\label{eq:adaptive_color_representation}
    \mathbf{K}_{\mathbf{g}} = \vert\vert \nabla_{sh} \mathcal{L}_{rad} \vert\vert_{2} = \sqrt{ \sum_{i=1}^{k}\left( \nabla_{h_{i}} \mathcal{L}_{rad} \right)^2 },
\end{equation}
where $k$ denotes the number of SH coefficients at the current order. The values of $k$ are 3, 12, 27, and 48 for orders 0 to 3, respectively. Importantly, we only accumulate derivatives corresponding to the current SH order, ensuring that $\mathbf{K}_{\mathbf{g}}$ faithfully reflects the adequacy of the current representation.

If $\mathbf{K}_{\mathbf{g}}$ exceeds a predefined threshold $\tau_{d}$, the SH order is increased to enhance expressive power. To balance efficiency and fidelity, we employ order-dependent thresholds: $\tau_{d} = 0.0001$ for upgrading from order 0 to 1 and order 1 to 2, and $\tau_{d} = 0.0002$ for order 2 to 3. This adaptive mechanism dynamically aligns parameter complexity with scene demands, reducing redundancy in smooth regions while preserving high-frequency details.

\subsubsection{Global: Task-decoupled Pruning}
After initialization and Gaussian separate operation, the scene is represented as a mixed Gaussian set $ \mathcal{G} = (\mathcal{G}^{m}, \mathcal{G}^{c}, \mathcal{G}^{n}) $.
In addition to pruning the essentially transparent Gaussian, we further follow prior works \cite{hanson2025pup, fan2024lightgaussian} to estimate the contribution of a Gaussian $\mathbf{g}$ by averaging its accumulated weights over all pixels $\mathbf{U}$ in which it participates:
\begin{equation}
\label{eq:task_decoupled_pruning}
\mathbf{W}_{\mathbf{g}} = \frac{1}{\left\|\mathbf{U}\right\|} \sum_{\mathbf{u} \in \mathbf{U}} \alpha_{\mathbf{g}}^{u} \mathbf{T}_{\mathbf{g}}^{u},
\end{equation}
where $\alpha_{\mathbf{g}}^{u}$ and $\mathbf{T}_{\mathbf{g}}^{u}$ denote the opacity and transmittance of Gaussian $\mathbf{g}$ on the pixel $\mathbf{u}$, respectively. 

Since we representation organizes the scene leveraging three subsets of Gaussians with different attributes (Common, Color-active and Normal-active Gaussian), directly ranking the contribution values of all Gaussians across subsets becomes ambiguous, as each subset is optimized under different supervisions. To address this issue, we adopt a task-decoupled pruning strategy that evaluates and prunes redundant Gaussians within each subset, independently.
Specifically, the Gaussians in the subsets ($\mathcal{G}_{m}, \mathcal{G}_{c}, \mathcal{G}_{n}$) are first sorted according to their task-specific contribution values. After a given iteration, the lowest $\tau_{g}\%$ Gaussians in each subset are pruned separately. 

Compared with coupled pruning\cite{Dai2024GaussianSurfels, fan2024lightgaussian}, 
this design enables selective redundancy pruning within each task, while better preserving reconstruction fidelity.

\subsection{Optimization}
\label{sec:opt}
To guide the optimization of our approach, we use the ground truth RGB images and the pseudo normal maps obtained from a pretrained neural network from Omnidata \cite{eftekhar2021omnidata} as supervisions. 
The pseudo normal maps are leveraged for warming-up rather than as the final training objectives. This is the reason why we dynamically update its loss weight in the training process, which aligns with the motivation of GaussianSurfels \cite{Dai2024GaussianSurfels}.
Detailed training strategies and loss function designs will be described in the following.

\subsubsection{Training Details}
\label{sec:opt_training_details}
We divide the training process into two stages, the optimization of CoRe and the application of Gaussian management.

There are two branches in CoRe: the SDF branch $ \mathcal{B}_{v} $ and the Gaussian branch $ \mathcal{B}_{g} $. 
The Gaussian branch $ \mathcal{B}_{g} $ is first warmed-up for 15,000 iterations so that its rendered depth $ \mathbf{D}_{g} $ can be used by the SDF branch $ \mathcal{B}_{v} $ for fine-grained ray sampling, and $ \mathcal{B}_{g} $ is finally optimized for 15,000 iterations after the training process of the volume branch $ \mathcal{B}_{v} $. So, $ \mathcal{T}_{i} $ mentioned in Eq.\ref{eq:normal_update} is set to 15,000.
The confidence mechanism is active throughout the process. We have presented the implementation of how confidence mechanism and volume results can improve the performance of Gaussian branches in Sec.\ref{sec:core}.

For the training of the SDF branch $ \mathcal{B}_{v} $, we implement an update version of \cite{instant-nsr-pl} and follow the GSDF \cite{yu2024gsdf} to sample points in the interval $ \left[ \mathbf{D}_{g} - k\vert \sigma(\mathbf{D}_{g}) \vert, \mathbf{D}_{g} + k \vert \sigma(\mathbf{D}_{g}) \vert \right] $, where we set $ k = 3 $ for all scenes. The optimization iteration of the SDF branch is set to 30,000.

After completing the optimization of the CoRe, we have obtained the final geometry used for supervision. Then we use $\mathcal{B}_{g} $ without the runtime confidence mechanism as the backbone to perform our Gaussian management to achieve the final result.
The optimization iteration is set to 15,000 for all datasets. 
Since our proposed Gaussian management is plug-and-play, we only need to disable the confidence mechanism in $ \mathcal{B}_{g} $ and replace normal supervision with the result of CoRe. We will explain the densification strategy in Sec.\ref{sec:densification_strategies}.
The time consumption of full pipeline and each component of our framework is summarized in Tab.~\ref{tab:time_consumption}.

\subsubsection{Loss Functions}
We utilize the loss function to make our approach obtain high-fidelity appearance and geometry. 
The loss functions used in optimizing $ \mathcal{B}_{g} $ and $ \mathcal{B}_{v} $ are defined as $ \mathcal{L}_{gaussian} $ and $ \mathcal{L}_{volume} $ ,respectively. Therefore, we have the following.
\begin{multline}
    \mathcal{L} _{gaussian} = \mathcal{L}_{rad} + \mathcal{L}_{geo} + \mathcal{L}_{con}^{a} + \lambda_{curv}\mathcal{L}_{curv} \\ 
    + \lambda_{opac} \mathcal{L}_{opac} + \lambda_{mask}\mathcal{L}_{mask} + \mathcal{L}_{\mathbf{F}_{g}}, 
\end{multline}
where $ \mathcal{L}_{curv} $ is the curvature loss, $ \mathcal{L}_{opac} $ and $ \mathcal{L}_{mask} $ are the opacity loss for attributes $ \mathbf{o} $ and the rendered mask $ \mathbf{O} $ defined in \cite{Dai2024GaussianSurfels}. We set $ \lambda_{curv} = 0.005 $, $ \lambda_{opac} = 0.01 $, $ \lambda_{mask} = 0.01 $ for the BMVS dataset \cite{yao2020blendedmvs} and $ \lambda_{mask} = 1 $ for the DTU dataset \cite{aanaes2016large}, respectively. 
$ \lambda_{n} $ and $ \lambda_{s} $ defined in Eq. (7) change linearly, $ \lambda_{n} $ decays from 0.04 to 0.02, and $ \lambda_{s} $ increases from 0.01 to 0.11.
For $ \mathcal{B}_{v} $, we follow the loss design $ \mathcal{L}_{nsr} $ of the work \cite{instant-nsr-pl} and add the normal loss $ \mathcal{L}_{vol} $ and the confidence loss $ \mathcal{L}_{\mathbf{F}_{v}}$.
\begin{equation}
    \mathcal{L}_{volume} =\mathcal{L}_{nsr} + \lambda_{vol} \mathcal{L}_{cos}(\mathbf{N}_{v}, \mathbf{N}_{g}) + \lambda_{conf} \mathcal{L}_{\mathbf{F}_{v}},
\end{equation}
where we use $ \lambda_{vol} = 0.01 $ and $ \lambda_{conf} = 0.005 $ in all experiments.

When applying Gaussian management, we follow the loss design $ \mathcal{L}_{gaussian} $ and update $ \mathbf{N} $ in $ \mathcal{L}_{geo} $ to the normal map of CoRe.
Therefore, the final loss function used in Gaussian management is defined as follows.
\begin{multline}
\label{eq:loss_student}
    \mathcal{L}_{manage} = \mathcal{L}_{rad}^{s} + \mathcal{L}_{geo}^{s} + \mathcal{L}_{con}^{s} + \lambda_{curv}\mathcal{L}_{curv} \\ 
     + \lambda_{opac} \mathcal{L}_{opac} + \lambda_{mask}\mathcal{L}_{mask}, 
\end{multline}
where $\mathcal{L}_{rad}^{s}, \mathcal{L}_{geo}^{s}$ and $\mathcal{L}_{con}^{s}$ are loss functions of radiance, geometry, and consistency.
\begin{equation}
\mathcal{L}_{rad}^{s} = (1 - \lambda_{s})\mathcal{L}_1(\mathbf{C}_{g}^{s}, \mathbf{C}) + \lambda_{s}\mathcal{L}_{ssim}(\mathbf{C}_{g}^{s}, \mathbf{C}),
\end{equation}
\begin{equation}
\mathcal{L}_{geo}^{s} = \lambda_{n}^{s}\mathcal{L}_{cos}(\mathbf{N}_{g}^{s}, \mathbf{N}_{g}) \odot \mathbf{M},
\end{equation}
\begin{equation}
\mathcal{L}_{con}^{s} = \lambda_{s}\mathcal{L}_{cos}(\mathbf{N}_{g}^{s}, \mathcal{F}(\mathbf{D_{g}^{s}}))\odot\mathbf{M},
\end{equation}

Since the normal map of CoRe is more faithful than the estimated normals, we set $ \lambda_{n}^{s} = 1 $ to balance the weight of appearance and geometry. 

\subsubsection{Densification Strategies}
\label{sec:densification_strategies}
Since optimization starts with sparse SFM points or random point clouds, we follow the densification strategies used in \cite{Dai2024GaussianSurfels} to increase the number of Gaussians for better reconstruction when optimizing CoRe.

In addition to the basic clone and split operations, Gaussian management also includes Gaussian separate operation, adaptive color representation, and task-decoupled pruning. 
We define the criterion for these operations after a single rendering in Sec.\ref{sec:gausep} and Sec.\ref{sec:gaurep}. The Gaussian separate operation and adaptive color representation is performed every $N$ iterations, where $N$ is the number of images, while task-decoupled pruning is utilized every $ 750 $ iterations for more stable optimization. Our detailed optimization and densification algorithms of Gaussian management are summarized in Algorithm \ref{alg:optimization}.
	\begin{algorithm}[!ht]
		\caption{Gaussian Management And Optimization}
		\label{alg:optimization}
		\begin{algorithmic}
			\STATE $ps, Rs, Ss, os, cs, ns \gets$ \textsc{InitAttributes}() 
			\STATE $i \gets 0$	
			
			\WHILE{not converged}
			
			\STATE $V, T \gets$ \textsc{GetItemTrainingData}()	
			\STATE $I \gets$ \textsc{Rasterize}($ps$, $Rs$, $Ss$, $os$, $cs$, $ns$, $V$)
			
			\STATE $L \gets$ \textsc{Loss}$(I, T) $ 
			
			\STATE $ps$, $Rs$, $Ss$, $os$, $cs$, $ns$ $\gets$ Adam($\nabla L$) 
			
		      \IF{\textsc{IterateAllImages}($i$)}
            \FORALL{$(p, R, S, o, c, n)$ $\textbf{in}$ $(ps, Rs, Ss, os, cs, ns)$}
            \STATE $\tau_d$ $\gets$ \textsc{GetOneupThreshold}($c$) 
            \IF{$\mathbf{K} > \tau_d$} 
            \STATE \textsc{OneupGaussianDegree}()
            \ELSE
            \IF{$\mathbf{P}<0$} 
            \STATE \textsc{SeparateGaussian}()
            \ELSE
            \IF{$ \nabla_p L > \tau_p$} 
            \STATE \textsc{CloneandSplitGaussian}()
            \ENDIF
            \ENDIF
            \ENDIF
            \ENDFOR
            \ENDIF

            \IF{\textsc{IsContriPruneIteration}($i$)}
            \STATE ${W}_g$ $\gets$ \textsc{GetContributeThres}($\tau_g$) 
            \FORALL{$(p, R, S, o, c, n)$ $\textbf{in}$ $(ps, Rs, Ss, os, cs, ns)$}
            \STATE \textsc{\textsc{TaskdecoupledPruning}}(${W}_g$)
            \ENDFOR
            \ENDIF
			\STATE $i \gets i+1$
			\ENDWHILE
		\end{algorithmic}
	\end{algorithm}


\section{Experiments}
\label{sec:exp}
We perform a thorough evaluation of our approach. 
First, the reconstruction results are evaluated. Then, the effectiveness and compatibility of key components is subsequently analyzed through comprehensive ablation studies.

\begin{figure*}[htbp]
\centering
\includegraphics[width=0.85\textwidth]{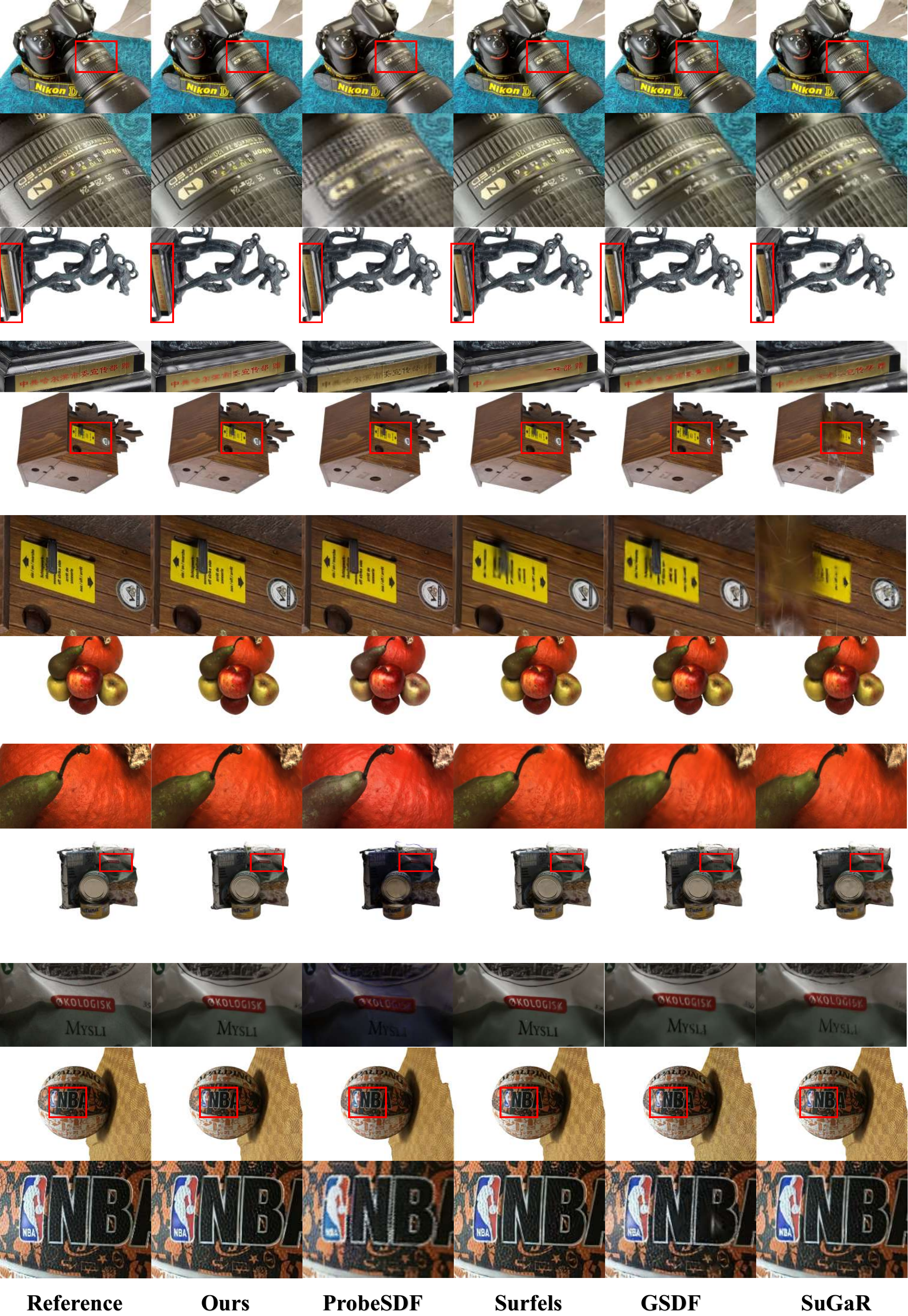}
\caption{\textbf{Qualitative Comparison of Appearance Reconstruction with SOTA methods on BMVS and DTU Datasets.}}
\label{fig:rendering_qualitative_comparison}
\end{figure*}

\begin{figure*}[htbp]
\centering
\includegraphics[width=0.98\textwidth]{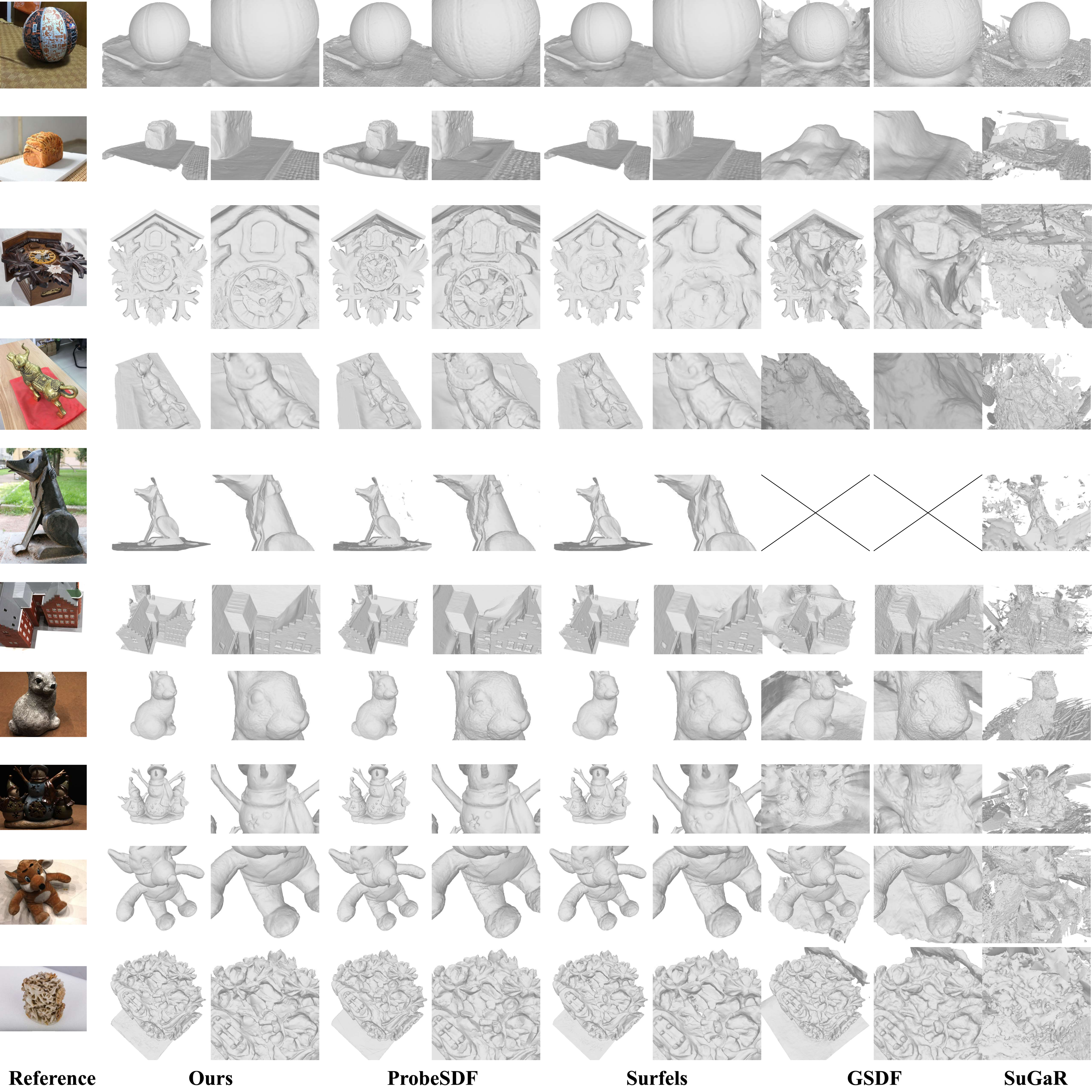}
\caption{\textbf{Qualitative Comparison of Surface Reconstruction with SOTA methods on BMVS and DTU Datasets.}}
\label{fig:mesh_qualitative_comparison}
\end{figure*}

\subsection{Implementation Details}
\subsubsection{Optimization}
We implement our approach on a GPU server with an i9-14900K CPU and a NVIDIA GeForce RTX 3090 GPU, using the Adam optimizer, PyTorch 2.1.2, and CUDA 11.8.
We capture all pixels of an image for Gaussian-related optimization and sample pixels for volume optimization.
The ADMM training procedure is utilized to let model start training at lower resolutions for warming-up.
The learning rate for the Gaussian position $\mathbf{p}$ is set to 1.6e-4 and decays exponentially to 1.6e-6, while the rates for other attributes—specifically $\mathbf{R}$, $\mathbf{S}$, $\mathbf{o}$, and $\mathbf{c}$—are set to 1.0e-3, 5.0e-3, 5.0e-2, and 2.5e-3, respectively.
Following \cite{Dai2024GaussianSurfels}, which demonstrates that random initialization of Gaussian positions and rotations does not degrade performance, we adopt this initialization strategy and leverage mask information to per-filter point clouds outside the scene on BMVS and DTU datasets. The SFM points are used to initialize the gaussian on MipNeRF360 and DeepBlending datasets.
across all experiments.
Moreover, the rasterization used in \cite{Dai2024GaussianSurfels} is re-wrote to perform Gaussian management. 
For the volume branch in CoRe, we use the AdamW optimizer and set the learning rate to 1e-2 for all network parameters.
The entire training process takes 1 $\sim$ 2 hours, depending on the complexity of the dataset and the resolution of the input image. 
Specifically, Gaussian management requires only 15 $\sim$ 20 minutes; \emph{CoRe}, along with the volume rendering module, accounts for the majority of the training time. 
We report a detailed time consumption of each component in our framework in Tab.~\ref{tab:time_consumption}.
\subsubsection{Surface Extraction}
\begin{figure*}[!htbp]
\centering
\includegraphics[width=0.98\textwidth]{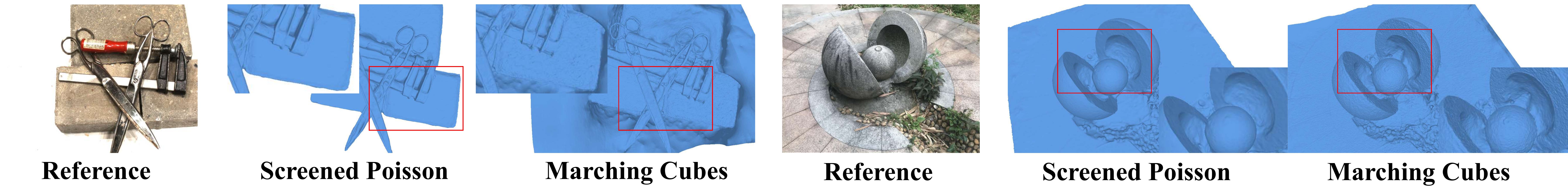}
\caption{\textbf{Comparison of surface extraction methods.} Screened Poisson reconstruction produces smoother and more watertight surfaces compared to Marching Cubes. 
Red boxes highlight regions where Screened Poisson better preserves fine geometric details.}
\label{fig:comparison_surface_extraction}
\end{figure*}

Both the Gaussian and volumetric branches in CoRe are capable of producing surface reconstructions. 
Instead of using the standard Marching Cubes algorithm, we adopt Screened Poisson Reconstruction utilized in \cite{kazhdan2013screened}, which has been shown to generate smoother and more watertight surfaces from multi-view rendered normal and depth maps. 
Leveraging the confidence mechanism, the Gaussian branch integrates the strengths of its own representation and volumetric rendering to produce more consistent and robust surface reconstructions.
Therefore, we use the Gaussian branch combined with Poisson reconstruction as our final surface extraction strategy.
In addition, we apply the volumetric cutting strategy proposed by \cite{Dai2024GaussianSurfels} to effectively remove floating artifacts. 
We present a comparison of surface extraction results using different methods for the two branches in Fig.\ref{fig:comparison_surface_extraction}.
After applying Gaussian management, three different types of Gaussian co-exist in the scene. 
Therefore, we leverage the precomputed volumetric cutting mask provided by CoRe to remove irrelevant regions to avoid ambiguity in the extraction process.
\subsubsection{Volumetric Cutting}
We follow the volumetric cutting strategy proposed in GaussianSurfel~\cite{Dai2024GaussianSurfels}, where the weighted opacity of each center of the voxel is used to approximate the integral of Gaussian weights and opacity within the corresponding intersection region. Voxels with low accumulated weighted opacity are regarded as unreliable and are subsequently removed from the initialized grid. During volumetric cutting, the grid resolution is set to $512^3$, and the accumulated weighted opacity threshold is set to 1. all Gaussian participate jointly in the voxel cutting process.
\subsection{Settings} 
\subsubsection{Baseline}
We select the current SOTA methods as the baseline, including 3DGS \cite{kerbl3Dgaussians}, SuGaR \cite{guedon2024sugar}, 2DGS \cite{huang20242d}, SURFELS \cite{Dai2024GaussianSurfels}, GSDF \cite{yu2024gsdf}, and ProbeSDF \cite{toussaint2025probesdf}. 
SuGaR, 2DGS and SURFELS are GS-based methods without any other representation, while GSDF and ProbeSDF are methods that use hybrid representation. 
GSDF hybridizes 3DGS with SDF, and ProbeSDF integrates voxel with radiance fields. 
The baseline comprehensively considers the SOTA methods in diverse representations. 
All comparisons use the original author's scripts, implementation, and hyper-parameters.

\subsubsection{Datasets}
Experiments are conducted on the BlenderMVS (BMVS) \cite{yao2020blendedmvs}, DTU \cite{aanaes2016large}, MipNeRF360 \cite{barron2022mip}, and Deep Blending \cite{hedman2018deep} public benchmark datasets to comprehensively evaluate our framework.
We follow the settings from SURFELS \cite{Dai2024GaussianSurfels} to select 18 and 15 scenes in the BMVS and DTU datasets, respectively, and we use $ 87.5\% (\frac{7}{8}) $ images for training and all images for evaluation for DTU dataset. The all images are used in training and evaluation for BMVS dataset.
The setting of the MipNeRF360 and Deep Blending datasets is derived from the GSDF \cite{yu2024gsdf}. 
All compared baselines adopt the same experimental settings for fairness.
For the BMVS and DTU datasets, we initialize the Gaussians with 100k randomly sampled points and the mask image is used to filter the background during training. For the DeepBlending and MipNeRF360 datasets, we initialize the Gaussians using SfM points generated by COLMAP and reconstruct the full scene. All compared methods adopt the same initialization strategy for fairness.

\subsubsection{Metrics}
Evaluation metrics include the Peak Signal-to-Noise Ratio (PSNR),
structural similarity (SSIM), Perceptual Image Patch Similarity (LPIPS), Chamfer distance (CD), F1-score, and the model size (Size, MB). 
We only count the parameters involved during inference and rendering for each method for the model size metric.
These metrics are comprehensive and include appearance and geometry.

\begin{figure}[!htbp]
    \centering
    \includegraphics[width=0.98\linewidth]{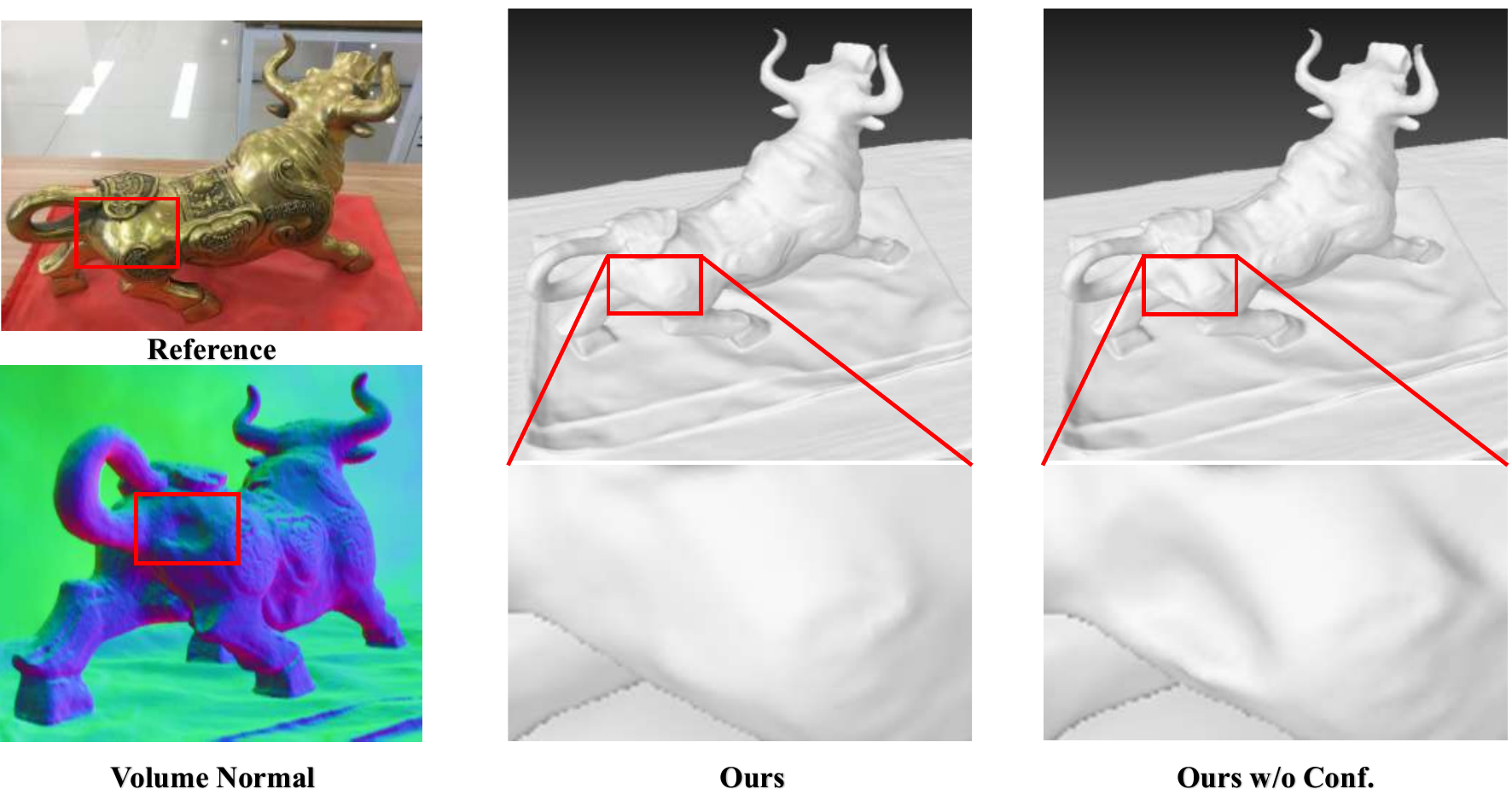}
    \caption{\textbf{Effect of the Confidence Mechanism.} It effectively prevents the under-reconstruction results of the SDF from being distilled into the Gaussian.}
    \label{fig:effect_of_confidence}
\end{figure}

\subsection{Quantitative and Qualitative Comparisons}
We compare our approach qualitatively and quantitatively.
First, the quantitative comparisons between our approach and other state-of-the-art methods on the BlenderMVS (BMVS) \cite{yao2020blendedmvs} and DTU \cite{aanaes2016large} datasets are reported in Tab. \ref{table:quality_comparison}. 
To further validate the generalization capability of our method, we additionally conduct quantitative evaluations on the MipNeRF360 \cite{barron2022mip} and Deep Blending \cite{hedman2018deep} datasets, as summarized in Tab.~\ref{table:extra_quality_comparison}. 
The results of 2DGS \cite{huang20242d} under the official settings are reported in the 2DGS$^{*}$ row.
Then we show the qualitative comparison on the reconstruction of \textbf{appearance} in Fig.~\ref{fig:rendering_qualitative_comparison} and \textbf{surface} in Fig.~\ref{fig:mesh_qualitative_comparison}. 
Compared with GS methods such as SuGaR \cite{guedon2024sugar}, 2DGS \cite{huang20242d} and GaussianSurfels \cite{Dai2024GaussianSurfels}, 
our approach demonstrates superior reconstruction performance for challenging scenes and fine details while maintaining a significantly lighting scene representation.
GSDF \cite{yu2024gsdf} proposes a hybrid representation composed of SDF and 3DGS, which achieves better PSNR and SSIM. However, the design of dual branch affects its robustness (Fig.~\ref{fig:mesh_qualitative_comparison}) and increases resource consumption (Tab. \ref{table:quality_comparison}). In contrast, our method achieves better results with only 2\% to 5\% parameters.
Moreover, ProbeSDF \cite{toussaint2025probesdf} prioritizes high-fidelity surface reconstruction while neglecting appearance reconstruction. It shows obvious color distortion (Okologisk mysli in Fig.~\ref{fig:rendering_qualitative_comparison}) of the novel view in the DTU dataset \cite{aanaes2016large}.
\begin{figure}[!htbp]
    \centering
    \includegraphics[width=0.98\linewidth]{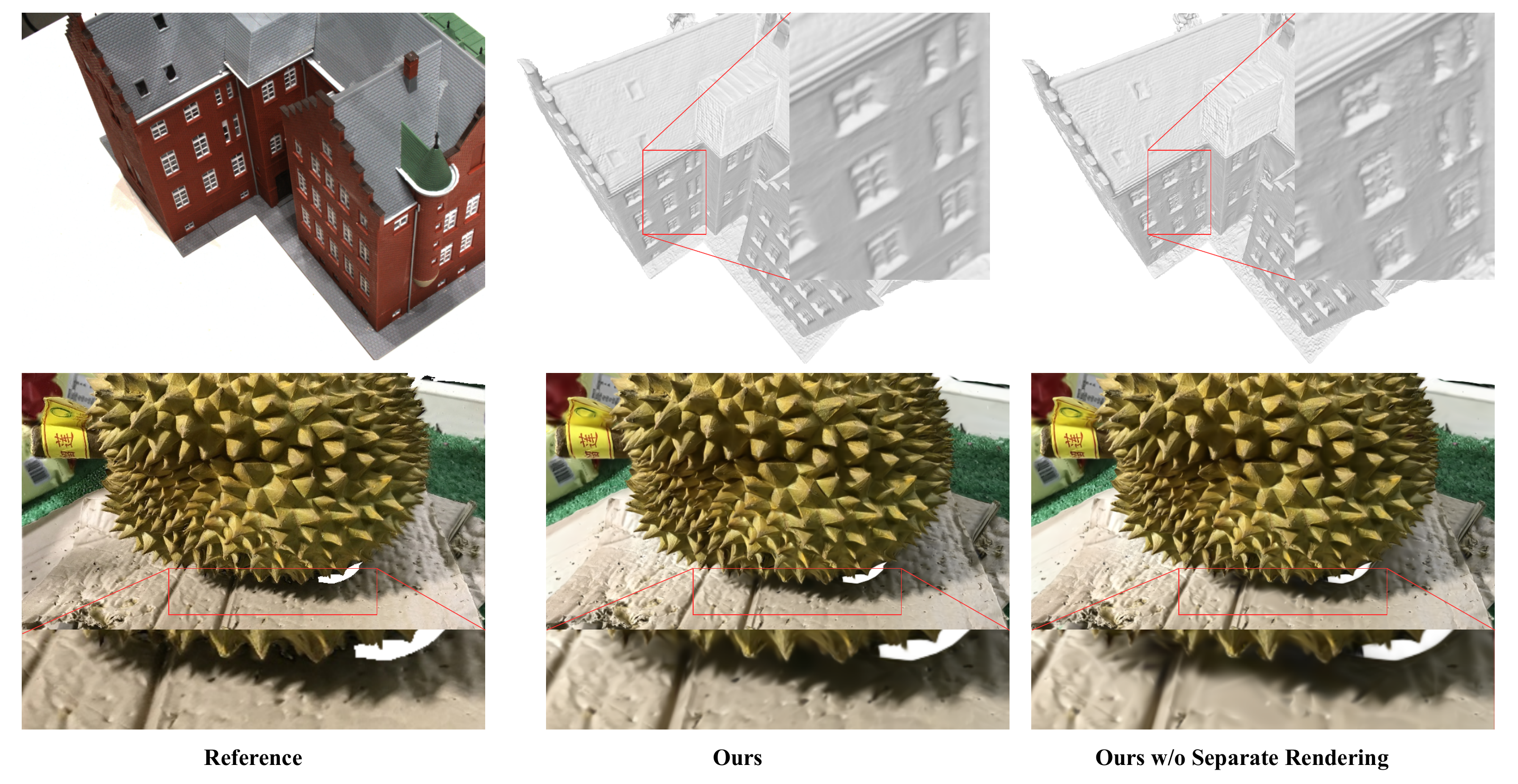}
    \caption{\textbf{Effect of Separate Rendering.} Separate rendering efficiently manages Gaussian attributes and avoids compromised performance, achieving better reconstruction of appearance (bottom) and surface (top).}
    \label{fig:effect_of_separate_rendering}
\end{figure}
\begin{table}[!htbp]
\caption{\textbf{Quantitative Comparison against baseline on BMVS and DTU Dataset.} SuGaR, GSDF, SURFELS and our approach initialize the primitives with \textbf{random} points. We run all baseline using official implementation for fair comparison. The official results of 2DGS under different settings are reported in 2DGS$^{*}$ row.}
\label{table:quality_comparison}
\centering
\setlength{\tabcolsep}{2mm}{
\begin{tabular}{cc|ccccc}
\hline\hline
\multicolumn{1}{c}{} & \multicolumn{1}{c | }{Method} & PSNR~$ \uparrow $ & SSIM~$ \uparrow $ & CD~$ \downarrow $ & F1~$ \uparrow $ & Size~$ \downarrow $\\
\hline
\multirow{7}{*}{\rotatebox{90}{\textbf{BMVS}}} & \multicolumn{1}{c | }{SuGaR} & 29.21 & 0.907 & 8.71 & 0.6317 & 222.2 \\
& \multicolumn{1}{c | }{2DGS} & 28.07 & 0.901 & 4.66 & 0.6646 & 98.82 \\
& \multicolumn{1}{c | }{2DGS$^{*}$} & - & - & - & - & - \\
& \multicolumn{1}{c | }{SURFELS} & 29.63 & \underline{0.932} & \underline{2.27} & \underline{0.8959} & 86.66 \\
& \multicolumn{1}{c | }{GSDF} & \underline{30.19} & 0.908 & 4.80 & 0.6380 & 811.8 \\
& \multicolumn{1}{c | }{ProbeSDF} & 29.90 & 0.885 & \textbf{2.24} & 0.8892 & \underline{50.00} \\
\cline{2-7}
& \multicolumn{1}{c | }{\textbf{Ours}} & \textbf{31.54} & \textbf{0.951} & 2.67 & \textbf{0.8968} & \textbf{45.14} \\
\hline
\multirow{7}{*}{\rotatebox{90}{\textbf{DTU}}} & SuGaR & 31.60 & 0.922 & 2.04 & 0.1580 & 259.5 \\
& 2DGS & 31.42 & 0.916 & 4.14 & 0.1371 & 48.17 \\
& \multicolumn{1}{c | }{2DGS$^{*}$} & 33.39 & 0.947 & 0.80 & - & - \\
& SURFELS & 33.21 & \underline{0.949} & 0.88 & 0.7873 & \underline{41.69} \\
& GSDF & \underline{33.66} & 0.949 & 0.80 & 0.6303 & 838.7 \\
& ProbeSDF & 33.49 & 0.944 & \textbf{0.66} & \textbf{0.8835} & 57.78 \\
\cline{2-7}
& \textbf{Ours} & \textbf{34.85} & \textbf{0.956} & \underline{0.80} & \underline{0.8392} & \textbf{17.44} \\
\hline\hline
\end{tabular}}
\end{table}
\begin{table}[!htbp]
\caption{\textbf{Quantitative Rendering comparison against baseline on Mip-NeRF360 and Deep Blending Dataset.} 3DGS, SuGaR, 2DGS, GaussianSurfel, GSDF and our approach initialize the primitives with the COLMAP sparse points.}
\label{table:extra_quality_comparison}
\centering
\setlength{\tabcolsep}{0.5mm}{
\begin{tabular}{c|ccc|ccc}
\hline\hline
Dataset & \multicolumn{3}{c|}{Mip-NeRF360} & \multicolumn{3}{c}{Deep Blending} \\
\hline
Method \& Metrics & PSNR~$\uparrow$ & SSIM~$\uparrow$ & LPIPS~$ \downarrow $ & PSNR~$\uparrow$ & SSIM~$\uparrow$ & LPIPS~$ \downarrow $ \\
\hline
3DGS & \underline{28.89} & 0.857 & 0.209 & 29.44 & 0.899 & 0.248 \\
SuGaR & 27.40 & 0.817 & 0.260 & 29.53 & 0.895 & 0.265 \\
2DGS & 28.08 & 0.843 & 0.240 & 29.27 & 0.897 & 0.261\\
GaussianSurfel & 28.71 & 0.860 & 0.196 & 27.22 & 0.857 & 0.364 \\
GSDF & \textbf{29.38} & \underline{0.865} & \underline{0.185} & \textbf{30.38} & \textbf{0.909} & \underline{0.223} \\
\hline
Ours & 28.74 & \textbf{0.888} & \textbf{0.147} & \textbf{30.51} & \underline{0.903} & \textbf{0.221} \\
\hline\hline
\end{tabular}}
\end{table}
\begin{table}[!htbp]
\caption{\textbf{Compatibility of Gaussian management (GM).} We replace CoRe with other works such as GSDF \cite{yu2024gsdf} and SURFELS \cite{Dai2024GaussianSurfels} to demonstrate its compatibility and the importance of high-quality supervision of geometry used for management.}
\label{table:vs}
\centering
\setlength{\tabcolsep}{2mm}{
\begin{tabular}{cc|cccc}
\hline\hline
\multicolumn{1}{c}{} & \multicolumn{1}{c | }{Setting} & PSNR~$ \uparrow $ & SSIM~$ \uparrow $ & CD~$ \downarrow $ & Size~$ \downarrow $ \\
\hline
\multirow{6}{*}{\rotatebox{90}{\textbf{BMVS}}} & GSDF & \textbf{30.19} & 0.908 & 4.80 & 811.80 \\
& GSDF w/ GM & 29.75 & \textbf{0.932} & \textbf{4.07} & \textbf{43.08} \\
\cline{2-6}
& SURFELS & 29.63 & 0.932 & \textbf{2.35} & 86.66 \\
& SURFELS w/ GM & \textbf{31.46} & \textbf{0.948} & 2.93 & \textbf{43.82} \\
\cline{2-6}
& CoRe & 31.33 & \textbf{0.960} & \textbf{2.24} & 220.0 \\
& CoRe w/ GM(ours) & \textbf{31.54} & 0.951 & 2.67 & \textbf{45.14} \\ 
\hline
\multirow{6}{*}{\rotatebox{90}{\textbf{DTU}}} & GSDF & 33.66 & 0.949 & 1.91 & 838.7 \\
& GSDF w/ GM & \textbf{34.85} & \textbf{0.954} & \textbf{1.19} & \textbf{17.19} \\
\cline{2-6}
& SURFELS & 33.21 & 0.949 & \textbf{0.90} & 41.69 \\
& SURFELS w/ GM & \textbf{34.83} & \textbf{0.956} & 0.94 & \textbf{17.69} \\
\cline{2-6}
& CoRe & 34.61 & \textbf{0.958} & 0.83 & 118.3 \\
& CoRe w/ GM(ours) & \textbf{34.85} & 0.956 & \textbf{0.80} & \textbf{17.44} \\ 
\hline\hline
\end{tabular}
}
\end{table}
\begin{figure}[!htbp]
    \centering
    \includegraphics[width=0.98\linewidth]{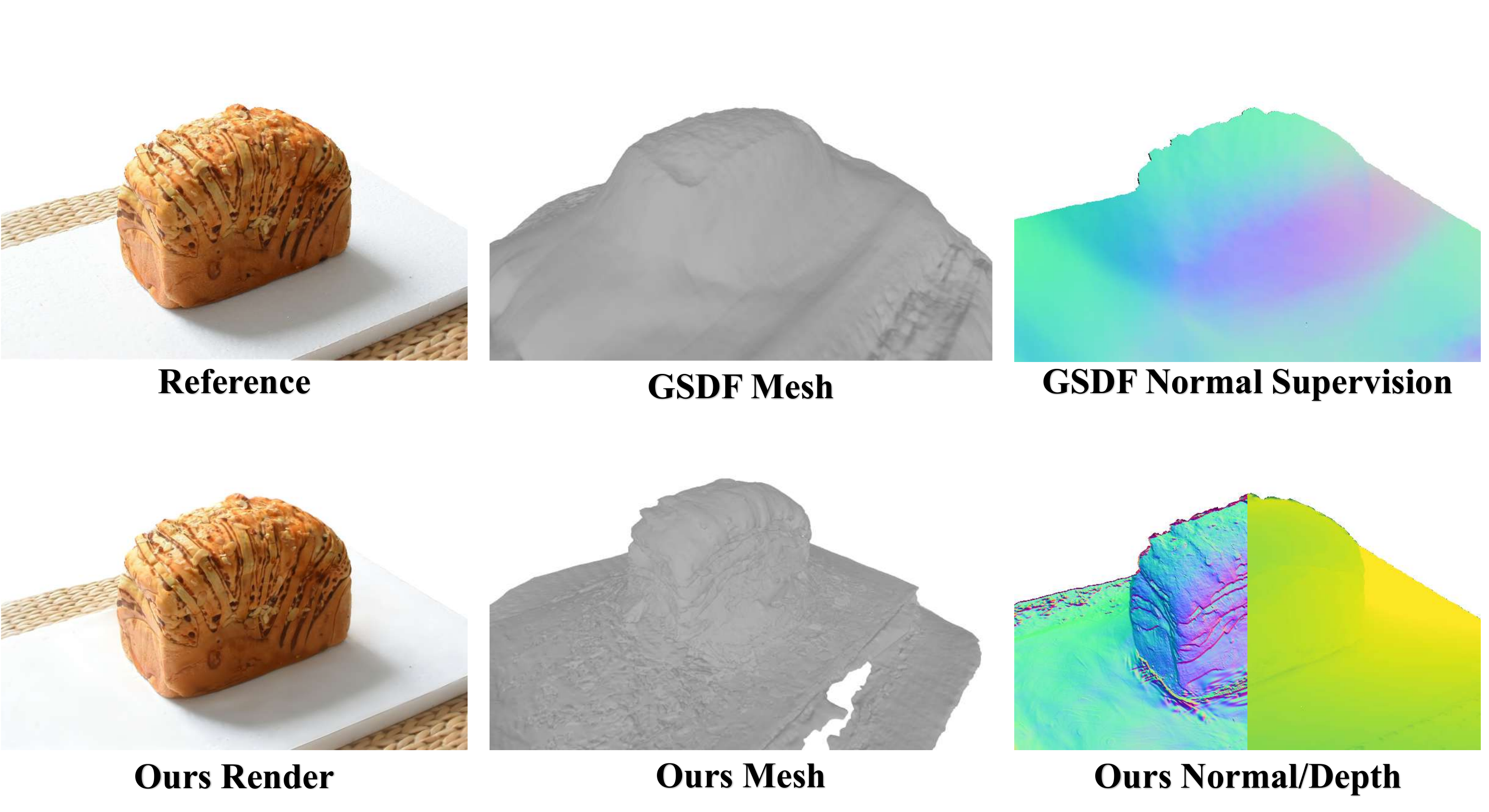}
    \caption{\textbf{Failure Cases.} When the geometry supervision used for management fails (using failure cases of GSDF to replace CoRe as an illustration), our method obtains poor surface reconstruction results.}
    \label{fig:failture_cases}
\end{figure}

\subsection{Ablation Studies} 
To verify the effectiveness of our design choices, we performed a series of ablation experiments of various major components of our method.
In general, the experiment settings and corresponding analyses are summarized as: 
\subsubsection{Confidence Mechanism(Conf.)}
We evaluate the effect of the confidence mechanism and present the result in the Tab. \ref{table:ab} (w/o Conf.). Using the confidence mechanism, our method obtains better geometry and further mapping this advantage to appearance. 
Fig. \ref{fig:effect_of_confidence} demonstrates the effectiveness of the confidence mechanism in suppressing noise signals.
\begin{table}[!htbp]
\caption{\textbf{Percentage of Gaussians across subsets and SH orders.} Values are in percentages (PCT, \%). “Color GS” and “Normal GS” denote color-active and normal-active Gaussians, and “D.~$X$ GS” indicates SH order $X$. Color-active Gaussians dominate, followed by the common set, while normal-active ones are the fewest. SH degrees 0 and 3 dominate the distribution, whereas degrees 1 and 2 are rare.}
\label{table:percent}
\centering
\setlength{\tabcolsep}{2mm}{
\begin{tabular}{c|ccc|cccc}
\hline\hline
\multirow{2}{*}{PCT} & Common & Color & Normal & D. 0 & D. 1 & D. 2 & D. 3 \\
& GS & GS & GS & GS & GS & GS & GS \\
\hline
BMVS & \underline{29.88} & \textbf{57.31} & 12.81 & \underline{32.76} & 0.08 & 1.57 & \textbf{65.59} \\
\hline
DTU & \underline{38.45} & \textbf{47.25} & 14.30 & \textbf{53.35} & 0.11 & 2.12 & \underline{44.42} \\
\hline\hline
\end{tabular}}
\end{table}
\subsubsection{Separate Rendering(S.R.)}
We evaluate the effect of densification based on separate rendering. As an important operation for controlling the quantity of Gaussian, the task-decoupled prune will lose its effect when the separate rendering is removed. Therefore, we replace it with the contribution prune operation that is effective for all Gaussians to avoid ambiguous experiments. The results are shown in Tab. \ref{table:ab} (w/o S.R.) and Fig. \ref{fig:effect_of_separate_rendering}. 
\subsubsection{Different Color Representation}
We evaluate the effect of the adaptive color representation. Specifically, we design four sets of experiments explicitly setting the order of SHs to fixed numbers, corresponding to D \#(0, 1, 2, 3) in the Tab. \ref{table:ab}. 
The results demonstrate that our method achieves order-3 reconstruction quality with parameters size equivalent to order 1 or 2, whereas a fixed order-3 configuration requires approximately 1.5 to 2 times the parameters of our adaptive approach.
\subsubsection{Effect of SDF branch}
We evaluate the effect of the SDF branch in CoRe module. The SDF branch is introduced as auxiliary to augment the geometry of the Gaussian branch. Specifically, we remove the SDF branch in our method and only utilize the Gaussian branch to provide geometry supervision for the Gaussian management. The results are shown in Tab.~\ref{table:ab} (w/o SDF). 
Although the SDF branch introduces additional computational cost during training, we retain it because it provides more stable geometry supervision for Gaussian optimization, as shown in Tab.~\ref{table:ab} (w/o SDF). The visualization in Fig.~\ref{fig:surfel_fault} further illustrate the robustness of the SDF branch in specific complex situation. Moreover, the SDF branch is only used during training and is completely discarded during inference, thus introducing no additional rendering overhead.
\subsubsection{Compatibility of Gaussian Management (GM)}
To demonstrate the compatibility of management that consists of GauSep and GauRep, we use different methods to replace CoRe, then report their re-run and updated results in the Tab. \ref{table:vs}. Management avoids redundancy while ensuring high-fidelity reconstruction. 
The results indicate that more accurate normal supervision further improves the final reconstruction quality and since our interact at the final supervision stage and manage the individual Gaussian primitive, the proposed Gaussian Management framework can be easily integrated into most existing 3DGS frameworks, demonstrating strong adaptability and compatibility.
Notably, we use their results to supervise the management. Therefore, we re-run and report the results of GSDF \cite{yu2024gsdf} and SURFELS \cite{Dai2024GaussianSurfels} in Tab. \ref{table:vs} for the sake of fairness.
\subsubsection{Prior Normal Supervision derived from Omnidata~\cite{eftekhar2021omnidata}}
We evaluate the effect of the prior normal supervision derived from OmniData~\cite{eftekhar2021omnidata}. The pseudo-normal maps generated by OmniData are utilized only during the early stage of Gaussian training to provide additional geometric supervision. To analyze their contribution, we remove the corresponding normal supervision loss during training. The experimental results are shown in Tab.~\ref{table:ab} (w/o Omn.).
The results indicate that the prior normal provides more geometric guidance during the early optimization stage, which further improves the final reconstruction quality. Nevertheless, even without OmniData supervision, the Gaussian branch is still able to reconstruct the scene geometry reasonably well due to the geometric representation capability of Gaussian primitives and the consistency constraints in our framework. This demonstrates that the effectiveness of Gaussian Management does not solely rely on external normal priors and exhibits good robustness under weaker geometric supervision.
\begin{table}[!htbp]
\caption{\textbf{Ablation Studies on BMVS \cite{yao2020blendedmvs} and DTU \cite{aanaes2016large} dataset of Key Component in our work.} We first evaluate the effectiveness of the confidence mechanism (w/o Conf.), then evaluate the effectiveness of separate rendering (w/o S.R.), and show experimental results under the different settings of color representation to prove the effectiveness of adaptive color representation (w/ D. \#0 - \#3). Finally, the effect of additional normal supervision is evaluated by remove the corresponding loss.}
\label{table:ab}
\centering
\setlength{\tabcolsep}{2mm}{
\begin{tabular}{cc|cccc}
\hline\hline
\multicolumn{1}{c}{} & \multicolumn{1}{c | }{Setting} & PSNR~$ \uparrow $ & SSIM~$ \uparrow $ & CD~$ \downarrow $ & Size~$ \downarrow $ \\
\hline
\multirow{8}{*}{\rotatebox{90}{\textbf{BMVS}}} & Ours w/o Conf. & 31.23 & 0.948 & 3.25 & 45.82 \\
\cline{2-6}
& Ours w/o S.R. & 30.52 & 0.939 & 2.69 & 49.51 \\
\cline{2-6}
& Ours w/ D.\#0 & 29.64 & 0.931 & 3.61 & 17.22 \\
& Ours w/ D.\#1 & 30.15 & 0.938 & 3.28 & 27.26 \\
& Ours w/ D.\#2 & 31.03 & 0.946 & 2.93 & 45.67 \\
& Ours w/ D.\#3 & 31.78 & 0.950 & 2.77 & 69.40 \\
\cline{2-6}
& Ours w/o SDF & 31.46 & 0.949 & 2.81 & 43.88 \\
\cline{2-6}
& Ours w/o Omn. & 30.96 & 0.948 & 2.65 & 63.35 \\
\cline{2-6}
& Ours & 31.54 & 0.951 & 2.67 & 45.14 \\
\hline
\multirow{8}{*}{\rotatebox{90}{\textbf{DTU}}} & Ours w/o Conf. & 34.76 & 0.955 & 0.87 & 17.04 \\
\cline{2-6}
& Ours w/o S.R. & 34.36 & 0.953 & 0.84 & 18.36 \\
\cline{2-6}
& Ours w/ D.\#0 & 33.91 & 0.949 & 1.11 & 8.98 \\
& Ours w/ D.\#1 & 34.18 & 0.952 & 0.95 & 13.99 \\
& Ours w/ D.\#2 & 34.70 & 0.956 & 0.87 & 22.67 \\
& Ours w/ D.\#3 & 35.22 & 0.958 & 0.84 & 34.35 \\
\cline{2-6}
& Ours w/o SDF & 34.83 & 0.956 & 0.92 & 18.51 \\
\cline{2-6}
& Ours w/o Omn. & 33.07 & 0.953 & 0.95 & 22.78 \\
\cline{2-6}
& Ours & 34.85 & 0.956 & 0.80 & 17.44 \\
\hline\hline
\end{tabular}
}
\end{table}
\begin{table*}[!htbp]
\caption{\textbf{The time consumption of our full method and each components in our method.} In the table, Gaussian and SDF represent the Gaussian branch and SDF branch in the CoRe, respectively. Although time consumption is concentrated in the training of the SDF branch, it only used in training process and not affect rendering speed.}
\label{tab:time_consumption}
\centering
\begin{tabular}{c|c|c|c|c|c|c|c|c}
\hline\hline
\multirow{2}{*}{Methods} & \multirow{2}{*}{Full} & \multirow{2}{*}{CoRe} & \multicolumn{2}{c|}{Gaussian Branch} & \multicolumn{2}{c|}{SDF Branch} & \multicolumn{2}{c}{Gaussian Management} \\
& & & Train & Render & Train & Render & Train & Render \\
\hline
Total Time & 95min23s & 81min53s & 23min50s & - & 58min03s & - & 13min30s & - \\
\hline
Iterations & 75k & 60k & 30k & - & 30k & - & 15k & - \\
\hline
Speed (Iter./s) & 13.11 & 12.21 & 20.98 & 69.42 & 8.61 & 0.04 & 18.52 & 99.67 \\
\hline\hline
\end{tabular}
\end{table*}
\subsection{Distribution and Percentage of each subsets of Gaussian}
\begin{figure}[!htbp]
    \centering
    \includegraphics[width=0.98\linewidth]{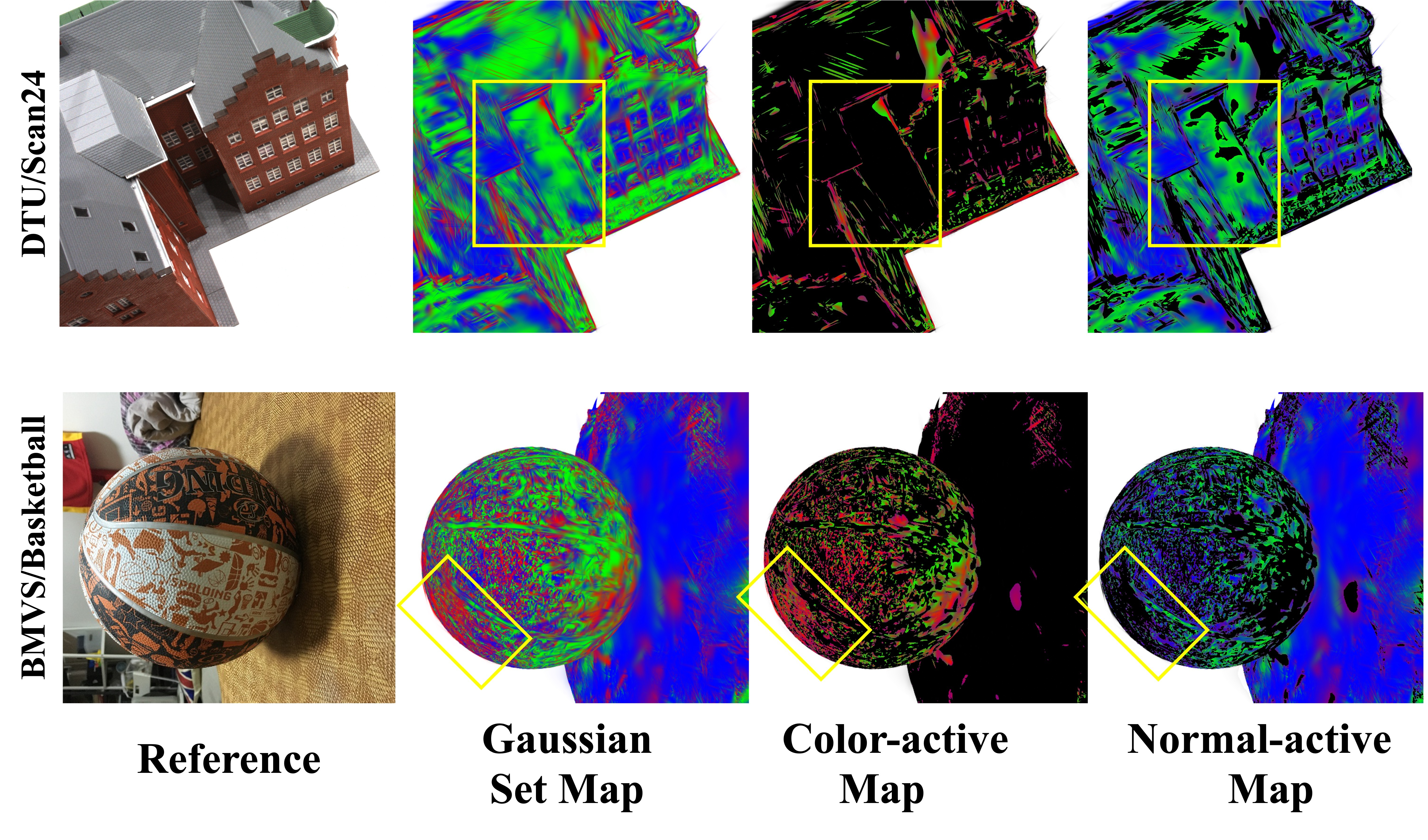}
    \caption{\textbf{Spatial Distributions of Gaussian.} The color of the Color-active, Normal-active and Common Gaussians is set to \textbf{red}, \textbf{blue} and \textbf{green}, respectively. Then we splat the Gaussian. In \textbf{DTU/Scan24}, regions highlighted in yellow contain mostly Normal-active Gaussians, reflecting easier appearance than geometry fitting. In \textbf{BMVS/Basketball}, highlighted regions are dominated by Color-active Gaussians, indicating more challenging appearance reconstruction. Overall, single-set Gaussians (color-active or normal-active) mainly appear in regions where the fitting difficulty between appearance and geometry differs, suggesting that Gaussians adaptively specialize in the attribute that is easier to optimize. We also visualize the corresponding point cloud on the right.}
    \label{fig:spatial_distribution}
\end{figure}

We represent the scene using three subsets of Gaussians: Color-active, Normal-active, and Common Gaussians. The spherical harmonics order (SHs) assigned to each Gaussian is also adaptive. Tab.~\ref{table:percent} reports the proportion of Gaussians in each subset together with the distribution of SH orders for the Color-active and Common subsets. In addition, Fig.~\ref{fig:spatial_distribution} visualizes the spatial distribution of these Gaussian subsets. We observe that in regions where the reconstruction difficulty of appearance and geometry is unbalanced, task-specific Gaussians (Color-active or Normal-active) are more likely to appear, enabling the representation to allocate capacity according to task complexity.
\begin{figure}[!htbp]
    \centering
    \includegraphics[width=0.98\linewidth]{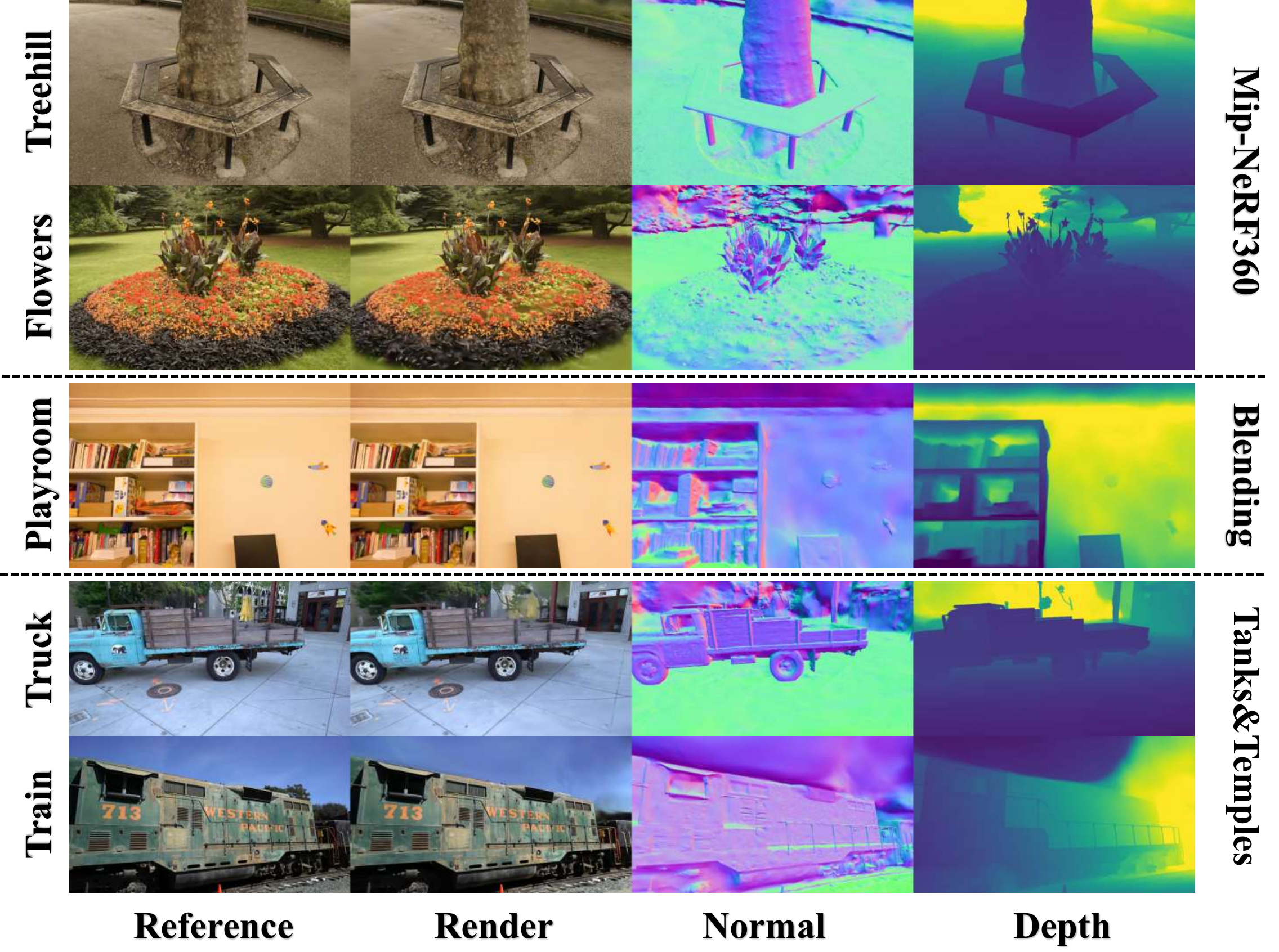}
    \caption{\textbf{Reconstruction results on unbounded scenes.} We evaluate our approach on five unbounded scenes selected from three different datasets. From top to bottom, the scenes include \emph{Treehill} and \emph{Flowers} from the Mip-NeRF360 dataset, \emph{Playroom} from the Deep Blending dataset, and \emph{Train} and \emph{Truck} from the Tanks$\And$Temples dataset. While our primary experiments are conducted on the BMVS and DTU datasets, which consist of bounded scenes or objects, the proposed approach also delivers high-quality performance on unbounded scenes. It is worth noting that, the sky regions are excluded from our reconstruction process. Therefore, the sky reconstruction yields the artificial results in the \emph{Train} scene.}
    \label{fig:unbounded_scene_reconstruction}
\end{figure}

\section{Discussion}
Our method is not without limitations. The first issue is the training time. On a \emph{single} RTX 3090 GPU, the training of a single scene takes 1 to 2 hours with the primary computational cost arising from the training of volume rendering. Second, our method shows a high dependence on high-precision geometric supervision, as shown in the Tab. \ref{table:vs} and Fig. \ref{fig:surfel_fault}. 
This dependency highlights an open question in geometry reconstruction. Within this constraint, our framework demonstrates how effective Gaussian management can achieve superior quality with minimal parameters. 
Finally, to ensure a more consistent evaluation in both qualitative and quantitative comparisons, we selected two object reconstruction datasets discussed in Sec.~\ref{sec:exp} from the currently available datasets, as they contain ground-truth geometry results. In addition, we also conducted experiments on several popular unbounded scene datasets\cite{barron2022mip, knapitsch2017tanks, hedman2018deep}, and the corresponding results are presented in Fig.~\ref{fig:unbounded_scene_reconstruction}. And our model-agnostic framework provides a foundation for integration with such approaches.

\section{Conclusion}
In this work, we propose a novel Gaussian management approach for high-fidelity scene reconstruction. By integrating a confidence-based surface reconstruction module, we enable reliable geometric supervision for surfel-based representations. Our separate rendering strategy allows Gaussians to selectively activate attributes, mitigating gradient conflicts, while the adaptive color representation dynamically adjusts expressiveness based on scene complexity. These designs jointly achieve a favorable result between appearance quality, geometry accuracy, and model size. Extensive experiments on standard benchmarks demonstrate that our approach outperforms state-of-the-art methods across multiple metrics and can be flexibly integrated into existing pipelines, highlighting its practical effectiveness and generality.

\section*{Acknowledgments}
This work was supported in part by the National Natural Science Foundation of China under Grant 62301278 and Grant 62371254.

\bibliographystyle{IEEEtran}
\bibliography{ref}

\begin{IEEEbiography}[{\includegraphics[width=1in,height=1.25in,clip,keepaspectratio]{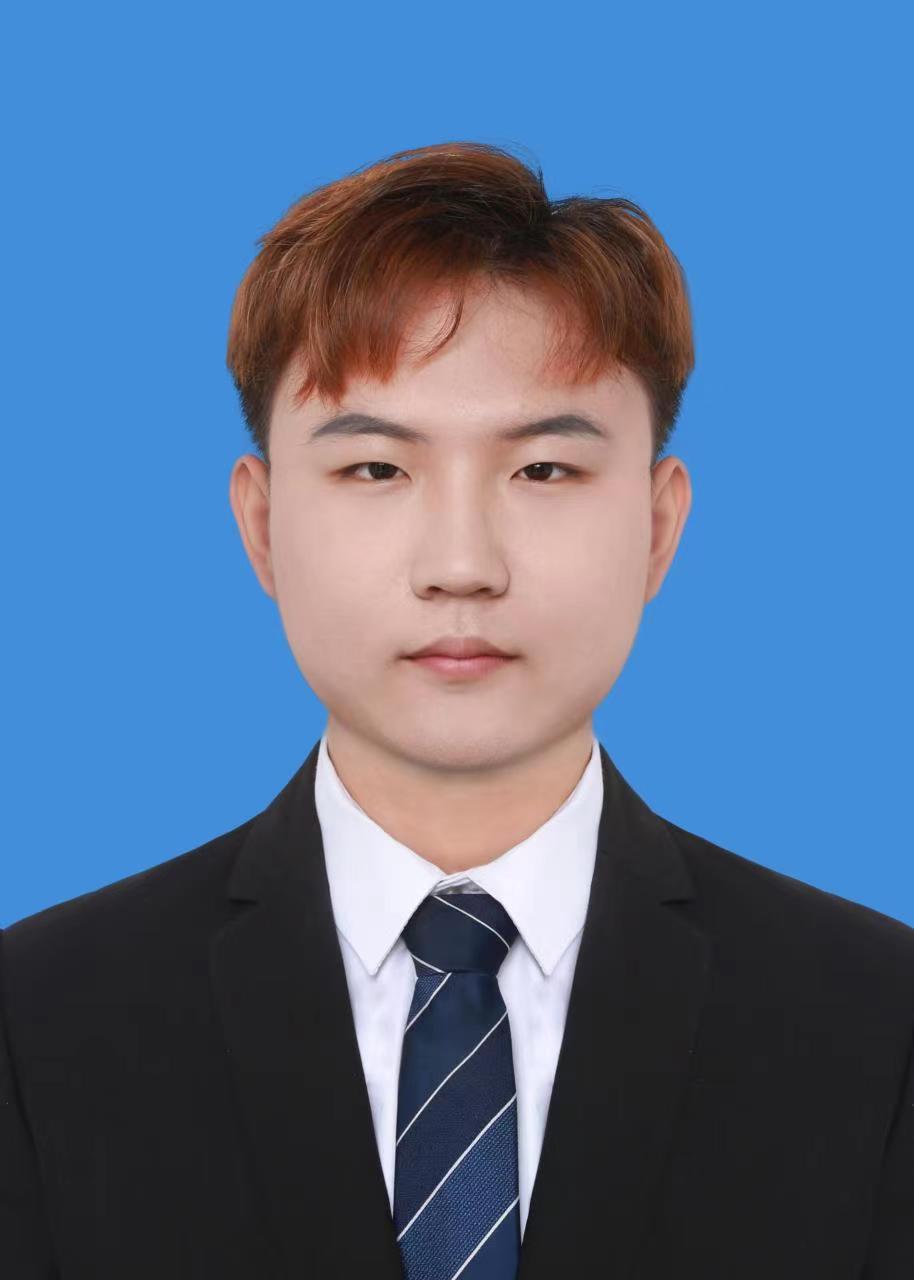}}]{Jiateng Liu} received the B.S. degree from Nanjing University of Posts and Telecommunications, Nanjing, China. He is currently pursuing the Ph.D. degree under the supervision of Prof. Hao Gao. His research interests include 3D computer vision and computer graphics.
\end{IEEEbiography}

\begin{IEEEbiography}[{\includegraphics[width=1in,height=1.25in,clip,keepaspectratio]{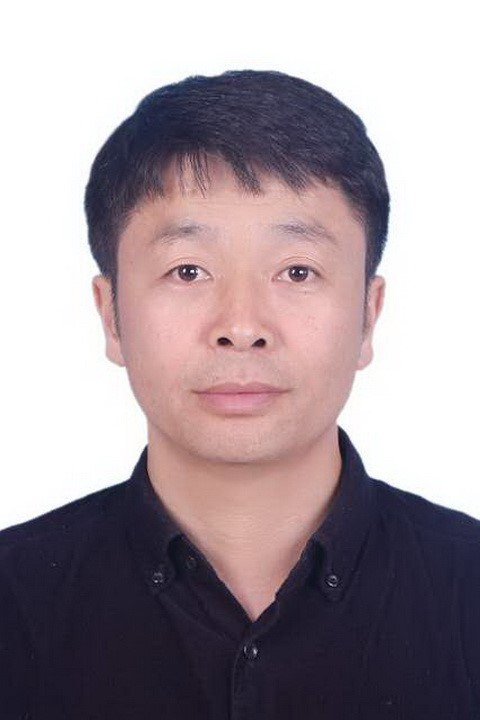}}]{Hao Gao} is currently a professor with the School of Automation, School of Artificial Intelligence, Nanjing University of Posts and Telecommunications, Nanjing, China. His research interests include artificial intelligence and computer vision, and he has authored or co-authored more than 50 international journal and conference papers. Prof. Gao was the Editorial Member and referee for many international journals.
\end{IEEEbiography}

\begin{IEEEbiography}[{\includegraphics[width=1in,height=1.25in,clip,keepaspectratio]{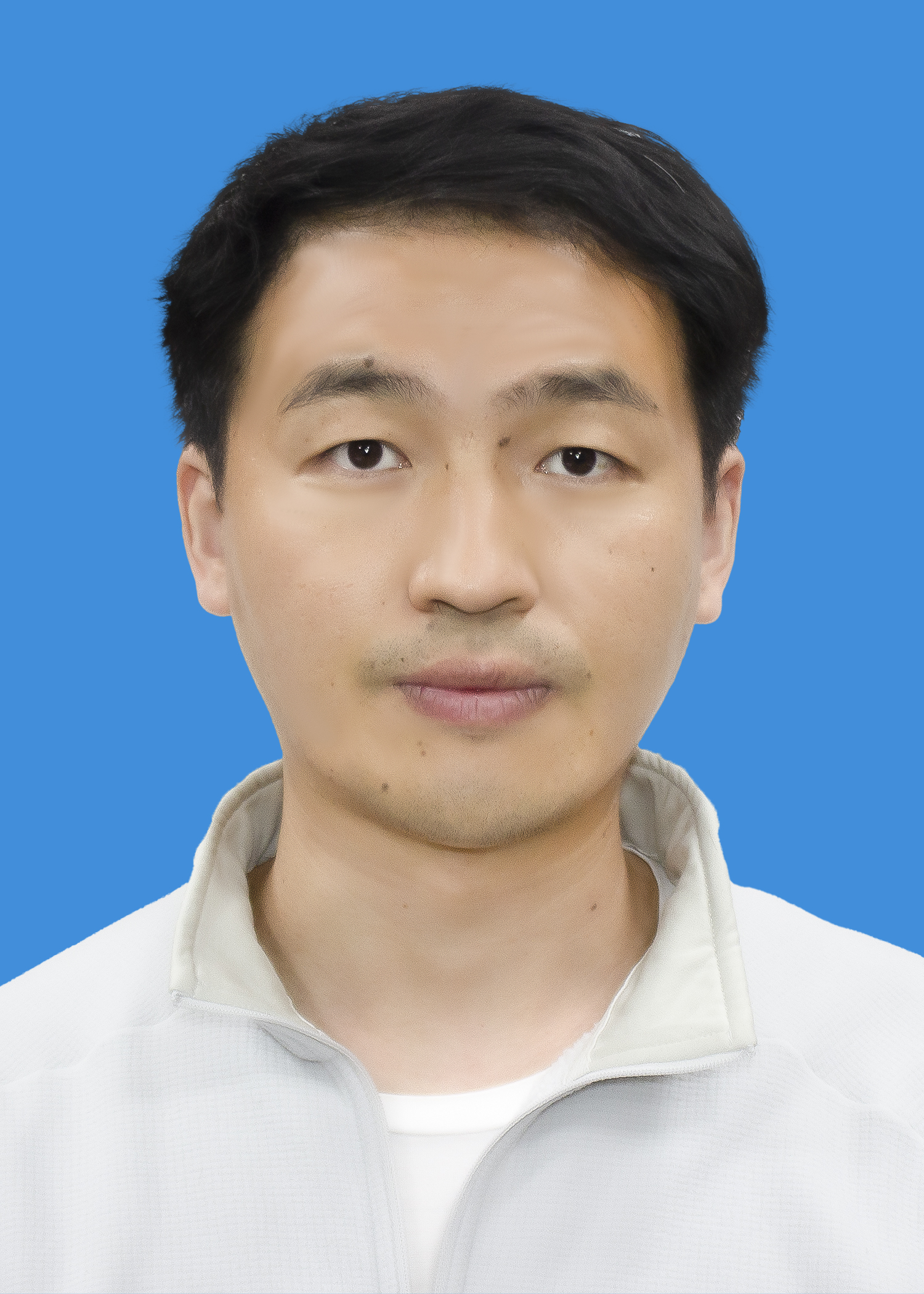}}]{Jiu-Cheng Xie} received his PhD degree in Computer and Information Science from the University of Macau in 2022, with a year of study at the Hong Kong Polytechnic University through a doctoral joint training program. He currently holds a lecturer position at the School of Automation and Artificial Intelligence, Nanjing University of Posts and Telecommunications, China. His research interests center on human-centric topics, including analysis, 3D reconstruction, animation, and their interactions with the environment.
\end{IEEEbiography}

\begin{IEEEbiography}[{\includegraphics[width=1in,height=1.25in,clip,keepaspectratio]{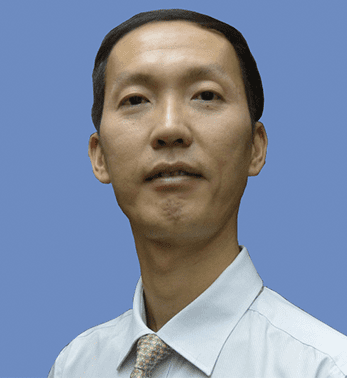}}]{Chi-Man Pun} received the B.Sc. and M.Sc. degrees in software engineering from the University of Macau in 1995 and 1998, respectively, and the Ph.D. degree in computer science and engineering from The Chinese University of Hong Kong in 2002. He was the Head of the Department of Computer and Information Science from 2014 to 2019. He is currently a Professor of computer and information science and in charge of the Image Processing and Pattern Recognition Laboratory, Faculty of Science and Technology, University of Macau. He has investigated many externally funded research projects as a PI, and has authored/coauthored more than 200 refereed papers in many top-tier journals and conferences. His research interests include image processing and pattern recognition; multimedia information security, forensic and privacy; adversarial machine learning and AI security. 
\end{IEEEbiography}

\begin{IEEEbiography}[{\includegraphics[width=1in,height=1.25in,clip,keepaspectratio]{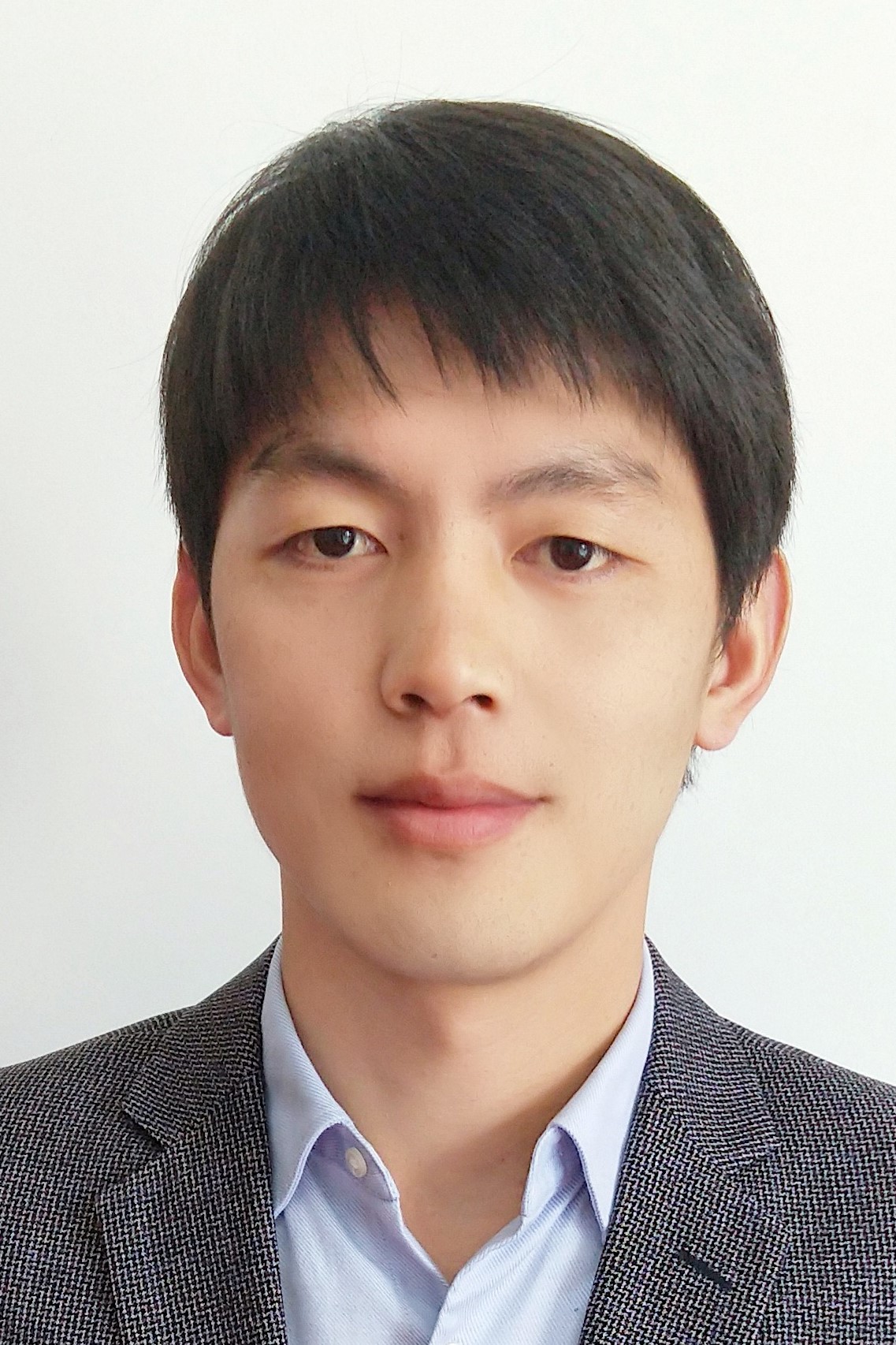}}]{Jian Xiong} received the Ph.D. degree from the University of Electronic Science and Technology of China, Chengdu, China, in 2015. In 2014, he was a Research Assistant at the Image and Video Processing Laboratory, The Chinese University of Hong Kong, Hong Kong. He is currently an Associate Professor with the School of Communications and Information Engineering, Nanjing University of Posts and Telecommunications, Nanjing, China. He has authored or co-authored more than 60 peerreviewed papers in international journals and conferences. He received the ICIP (International Conference on Image Processing) Most Influential Paper Award (2024) and the Best Paper Award from the International Conference on Advanced Hybrid Information Processing (2018). His research interests include image and video processing, point cloud compression, and computer vision.
\end{IEEEbiography}

\begin{IEEEbiography}[{\includegraphics[width=1in,height=1.25in,clip,keepaspectratio]{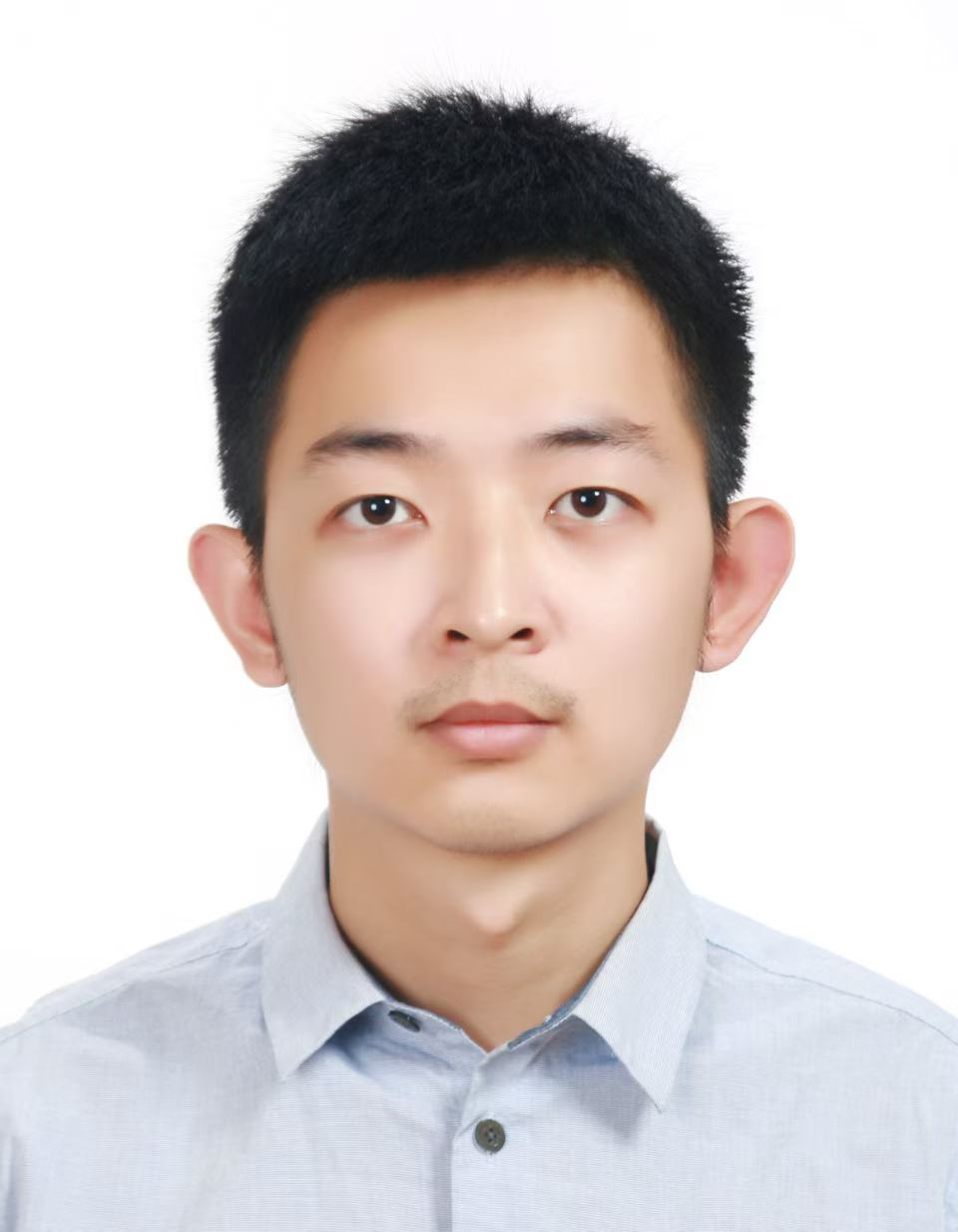}}]{Haolun Li} received the M.sc. degree in pattern recognition and intelligent system from the Nanjing University of Posts and Telecommunications, China, in 2020. He received his Ph.D. degree from the University of Macau in 2024. His current research interests include computer vision and human motion analysis.
\end{IEEEbiography}

\begin{IEEEbiography}[{\includegraphics[width=1in,height=1.25in,clip,keepaspectratio]{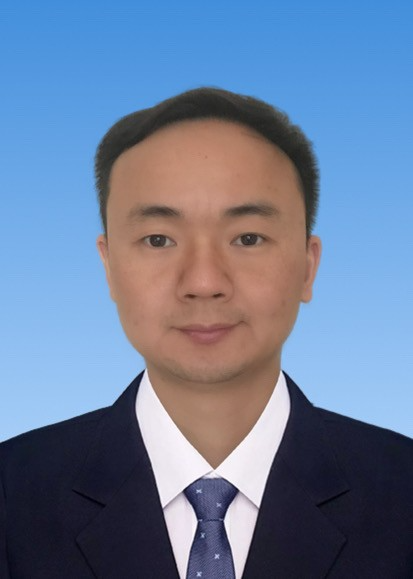}}]{Junxin Chen} (Senior Member, IEEE) is currently a Full Professor at the School of Software, Dalian Uni-versity of Technology, Dalian, China. He received the B. Sc. Degree in Communications Engineering and M.Sc. and Ph.D degrees in Communications and Information System, all from Northeastern Uni-versity in 2007, 2009 and 2016, respectively. From2019 to 2020, he was a Postdoc Research Fellow in University of Macau, under the UM Macau Talent Programme (Class A). He worked as an Assistant Professor and Associate Professor in the College of Medicine and Biological Information Engineering, Northeastern University,Shenyang,China, and is currently working as a Full Professor at the School of Software, Dalian University of Technology.
His research interests include ubiquitous healthcare, Internet of Medical Things, artificial intelligence. He has authored over 150+ scientific papers in international peer-reviewed journals and conferences, and has a H-index of 37and a total of 4200+ citations. He is an area editor of NeurIPS, associate editor of Expert Systems (Wiley), International Journal of Interactive Multimedia and Artificial Intelligence, the (Leading) Guest Editor of IEEE Journal of Biomedical and Health Informatics, etc. He has received about 25 awards from Mainland China and Macau, and has given more than 30 invited talks for universities and companies in China and Australia. He is awarded Xiaomi Young Talents Program under Xiaomi Foundation.
\end{IEEEbiography}

\begin{IEEEbiography}[{\includegraphics[width=1in,height=1.25in,clip,keepaspectratio]{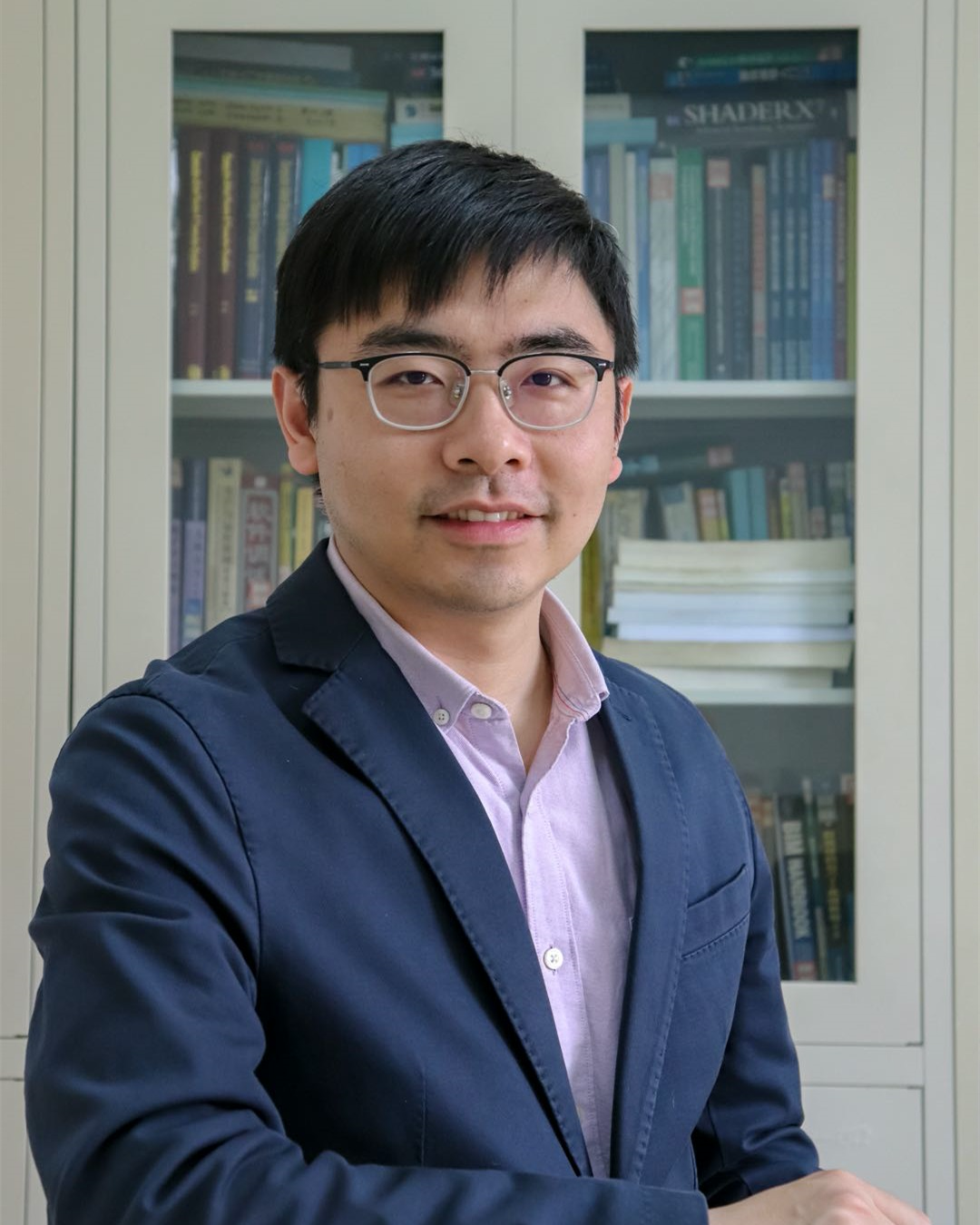}}]{Feng Xu} received the BS degree in physics from Tsinghua University, Beijing, China, in 2007, and the PhD degree in automation from Tsinghua University, Beijing, China, in 2012. He is currently an associate professor with the School of Software, Tsinghua University, China. His research interests include facial animation, performance capture, and 3D reconstruction.
\end{IEEEbiography}

\newpage

\vfill

\end{document}